\AtBeginDocument{%
  }

\documentclass[sigconf]{acmart}
\usepackage{multirow}
\usepackage{enumitem}
\copyrightyear{2025}
\acmYear{2025}
\setcopyright{acmlicensed}
\acmConference[MM '25] {Proceedings of the 33rd ACM International Conference on Multimedia}{October 27--31, 2025}{Dublin, Ireland.}
\acmBooktitle{Proceedings of the 33rd ACM International Conference on Multimedia (MM '25), October 27--31, 2025, Dublin, Ireland}
\acmISBN{979-8-4007-2035-2/2025/10}
\acmDOI{10.1145/3746027.3755726}

\begin{document}

\title{Beyond Emotion Recognition: A Multi-Turn Multimodal Emotion
Understanding and Reasoning Benchmark}


\author{Jinpeng Hu}
\affiliation{%
  \institution{Hefei University of Technology}
  \state{Hefei}
  \country{China}}
\email{jinpenghu@hfut.edu.cn}

\author{Hongchang Shi}
\affiliation{%
  \institution{Hefei University of Technology}
    \state{Hefei}
  \country{China}}
\email{2024170833@mail.hfut.edu.cn}

\author{Chongyuan Dai}
\affiliation{%
   \institution{Hefei University of Technology}
  \state{Hefei}
  \country{China}}
\email{taisungyun@mail.hfut.edu.cn}

\author{Zhuo Li}
\affiliation{%
  \institution{The Chinese University of Hong Kong, Shenzhen}
  \state{Shenzhen}
  \country{China}}
  \email{221019088@link.cuhk.edu.cn}

\author{Peipei Song}
\authornote{Corresponding authors.}
\affiliation{%
  \institution{University of Science and Technology of China}
  \state{Hefei}
  \country{China}}
\email{beta.songpp@gmail.com}

\author{Meng Wang}
\authornotemark[1]
\affiliation{%
   \institution{Hefei University of Technology}
   \institution{Institute of Artificial Intelligence (IAI), Hefei Comprehensive National Science Center}
  \state{Hefei}
  \country{China}}
\email{eric.mengwang@gmail.com}

\renewcommand{\shortauthors}{Jinpeng Hu et al.}


\begin{abstract}

Multimodal large language models (MLLMs) have been widely applied across various fields due to their powerful perceptual and reasoning capabilities. In the realm of psychology, these models hold promise for a deeper understanding of human emotions and behaviors. However, recent research primarily focuses on enhancing their emotion recognition abilities, leaving the substantial potential in emotion reasoning, which is crucial for improving the naturalness and effectiveness of human-machine interactions. Therefore, in this paper, we introduce a multi-turn multimodal emotion understanding and reasoning (MTMEUR) benchmark, which encompasses 1,451 video data from real-life scenarios, along with 5,101 progressive questions. These questions cover various aspects, including emotion recognition, potential causes of emotions, future action prediction, etc. Besides, we propose a multi-agent framework, where each agent specializes in a specific aspect, such as background context, character dynamics, and event details, to improve the system's reasoning capabilities. Furthermore, we conduct experiments with existing MLLMs and our agent-based method on the proposed benchmark, revealing that most models face significant challenges with this task.

\end{abstract} 

\ccsdesc[500]{Applied computing~Psychology}
\ccsdesc[300]{Information systems~Multimedia streaming}
\keywords{Emotion Reasoning, Emotion Recognition, Mutimodal}

\begin{abstract}

Multimodal large language models (MLLMs) have been widely applied across various fields due to their powerful perceptual and reasoning capabilities. In the realm of psychology, these models hold promise for a deeper understanding of human emotions and behaviors. However, recent research primarily focuses on enhancing their emotion recognition abilities, leaving the substantial potential in emotion reasoning, which is crucial for improving the naturalness and effectiveness of human-machine interactions. Therefore, in this paper, we introduce a multi-turn multimodal emotion understanding and reasoning (MTMEUR) benchmark, which encompasses 1,451 video data from real-life scenarios, along with 5,101 progressive questions. These questions cover various aspects, including emotion recognition, potential causes of emotions, future action prediction, etc. Besides, we propose a multi-agent framework, where each agent specializes in a specific aspect, such as background context, character dynamics, and event details, to improve the system's reasoning capabilities. Furthermore, we conduct experiments with existing MLLMs and our agent-based method on the proposed benchmark, revealing that most models face significant challenges with this task.

\end{abstract} 
\maketitle

\begin{figure*}[t]
\centering
    \includegraphics[width=\textwidth]{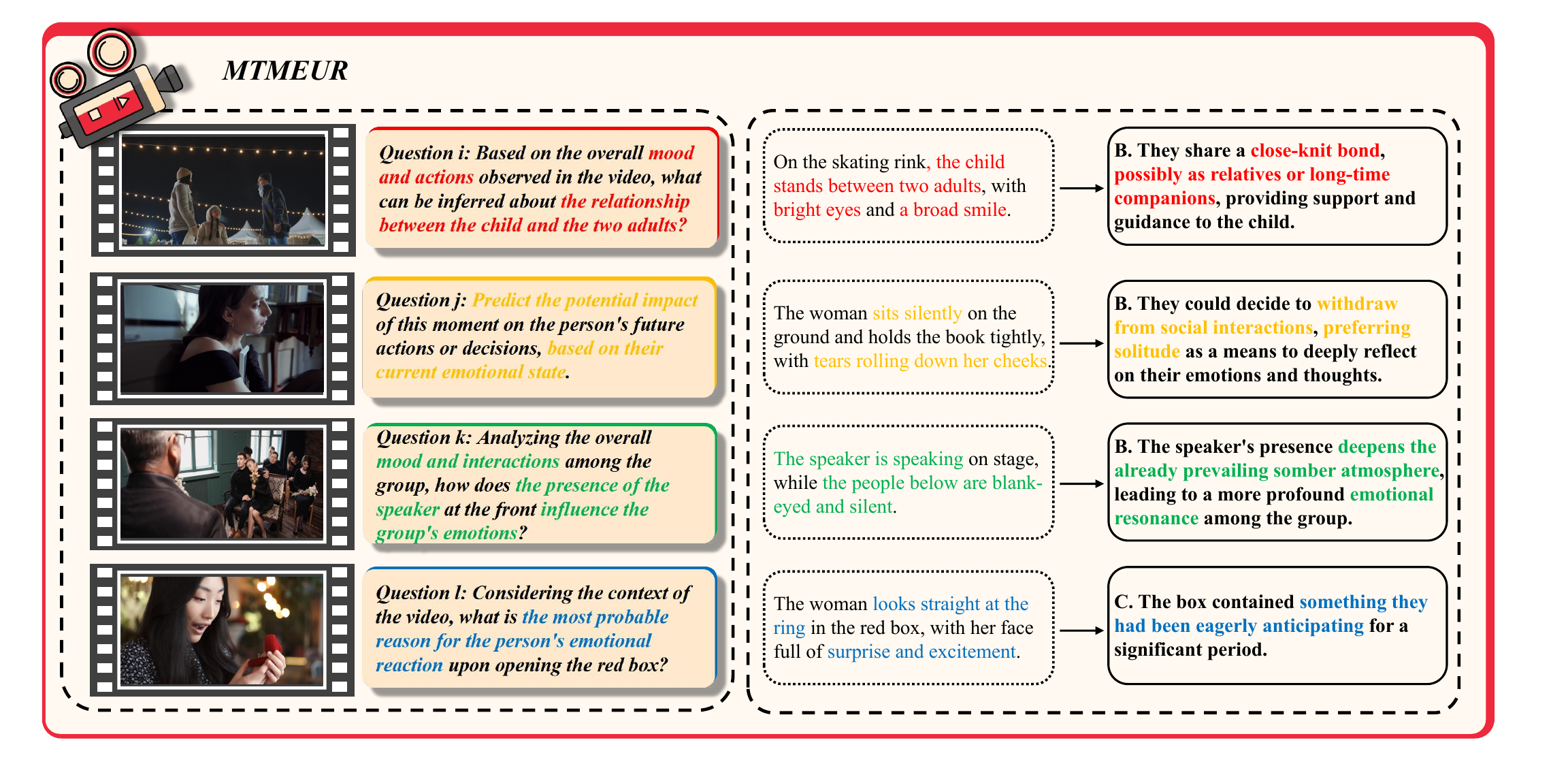}
    \caption{Examples from MTMEUR, presenting four questions along with their answers and the reasoning processes necessary for generating the correct responses.}
    \label{fig_intro}
\end{figure*}
\section{Introduction}
\label{sec:intro}

In recent years, multimodal large language models (MLLMs) have shown significant advancements across diverse fields due to their enhanced capabilities in multimodal understanding and reasoning\cite{li2024llava2,liu2024oryx}.
One of the ultimate goal of MLLMs is to improve human-machine interactions, enabling more seamless and intuitive communication.
Central to achieving this goal is the development of multimodal emotion understanding and reasoning, as a deeper understanding of human emotions allows for responses that are more adaptive and acceptable to users.
Therefore, modeling and evaluating emotional and psychological states have drawn substantial attention in recent years, and many studies have been proposed in this area\cite{gratch2014distress,mundnich2020tiles,coppersmith2015clpsych,turcan2019dreaddit,rashkin1811towards}.
For example, the CMU-MOSI dataset\cite{zadeh2016multimodal} provided video segments of utterances annotated for sentiment polarity and intensity, enabling researchers to assess model performance in multimodal sentiment analysis. 
Furthermore, the MELD dataset\cite{poria2019meld} offered video clips of multi-party conversations annotated with six different emotions, enabling researchers to evaluate model performance in emotion recognition. 
Moreover, the Social-IQ\cite{zadeh2019social} provided a resource for assessing social intelligence through question-answering tasks, enabling researchers to gauge model performance in understanding social and emotion cues.
MEmoR\cite{shen2020memor} introduced a multimodal emotion recognition dataset that integrates a reasoning process, leveraging supporting evidence to better recognize emotion.

Although these approaches have brought significant improvements, several issues cannot be appropriately considered.
First, most existing studies emphasize assessing sentiment intensity, polarity, and emotion types\cite{wollmer2013youtube,fang2022faf}.
However, they still fall short in capturing the complexity of emotion interactions and do not sufficiently explore the evolving relationships and transitions between emotions.
Second, most benchmarks lack detailed exploration of emotion-specific aspects, resulting in limitations in both scope and depth\cite{shen2020memor}.
For example, as shown in Figure \ref{fig_intro}, identifying potential causes of emotions or predicting future human responses within the video context is essential for a more comprehensive understanding of emotional and psychological states, while such elements are overlooked in datasets such as Social-IQ and Social-IQ2.

Therefore, in this paper, we introduce a multi-turn multimodal emotion understanding and reasoning (MTMEUR) benchmark, which consists of 1,451 carefully curated videos that cover a wide range of complex scenarios, along with 5,101 high-quality progressive questions, including emotion recognition, potential causes of emotions, future action prediction, etc.
Methodologically, we propose a novel multi-agent-based approach to improve the emotion reasoning capabilities of MLLMs by efficiently extracting and integrating distinct aspects of information.
This framework consists of four specialized agents, each concentrating on a unique aspect:  background context, character dynamics, event details, and decision-making processes.
Furthermore, we evaluate several state-of-the-art (SOTA) MLLMs and the proposed method on MTMEUR dataset, with accuracy ranging from 29.10\% (Videochatgpt) to 71.19\% (Qwen2-VL).
The experimental results also show that the proposed method is beneficial to improving emotion reasoning ability.
Our code is released at \url{https://github.com/MindIntLab-HFUT/MTMEUR}.

\section{Related works}
\label{Related works}
\subsection{Multimodal Large Language Models}
Natural Language Processing (NLP) methodologies have been extensively applied across diverse domains, encompassing tasks such as text summarization, information extraction, and text classification \cite{liu2019text, hu2021word,hu2022graph,hu2022hero,li2025prototype}.
Large Language models have revolutionized NLP tasks, achieving success across a broad spectrum of applications \cite{hu2024psycollm,li2025add,li2024self,goel2023llms}.
Foundational architectures like Transformer\cite{vaswani2017attention} and training paradigms like instruction tuning\cite{wei2021finetuned} have enabled models such as PaLM\cite{chowdhery2023palm}, and LLaMA\cite{touvron2023llama} to demonstrate remarkable language understanding capabilities.
Building upon these textual foundations, the emergence of MLLMs has extended these capabilities to visual domains.
The debut of GPT-4 (Vision) has attracted significant attention due to its groundbreaking multimodal understanding capabilities.
This success has sparked a wave of research into MLLMs. 
Early research in this area primarily focused on image-text understanding tasks\cite{li2024llava,yan2024list,zhang2023llavar,zhao2023svit}.
LLaVA\cite{liu2024visual} and MiniGPT-4\cite{zhu2023minigpt}, where image encoders were utilized to extract visual features and were then aligned with LLM-based encoders for further processing.
These models have paved the way for new approaches to multimodal content understanding and generation. 
As research has progressed, some models have begun to incorporate video modalities, exploring the vast potential of large models in video understanding. 
VideoChat\cite{li2023videochat} integrated video foundation models and LLM via a learnable neural interface.
Furthermore, Video-ChatGPT\cite{maaz2023video} used CLIP\cite{radford2021learning} to extract spatial and temporal features from videos, which are then integrated into LLM to enhance understanding and generating detailed conversations. 
\subsection{Emotion Recognition Dataset}

In recent years, researchers have made significant progress in the field of emotion understanding, including emotion recognition \cite{abdullah2021multimodal}, emotional video captioning \cite{song2024emotional, song2023emotion, song2023contextual} and so on.
Many benchmarks have been proposed to evaluate the model performance in terms of emotion recognition  \cite{xu2019multi,xu2022mcpr}.
For instance, CMU-MOSI \cite{zadeh2016multimodal}, CMU-MOSEAS \cite{zadeh2020cmu}, and Youtube \cite{morency2011towards} integrated text, and video modalities, and offered basic emotion annotations, which focused on predicting sentiment polarity and intensity within videos.
Furthermore, MELD \cite{poria2019meld}, MuSe-CaR \cite{stappen2021multimodal}, and IEMOCAP \cite{busso2008iemocap} contained videos of multi-party conversations, which further incorporated the context of interactions.
CMU-MOSEI\cite{zadeh2018multimodal} consists of numerous YouTube video clips, each containing at least one recognized emotion, further enriching the multimodal emotion analysis.
In addition, some recent studies \cite{zadeh2019social,siq2} focus on measuring emotion cognition and intelligence within social contexts through question answering.
Compared to existing benchmarks, the proposed MTMEUR focuses on the reasoning of characters in a video.
This task demands not only an advanced understanding of the visual and contextual elements within the video, but also necessitates the integration of sophisticated common sense knowledge.
MTMEUR challenges the model to simulate human-like responses within specific emotion contexts and scenarios, offering a better evaluation of its emotion reasoning and cognitive abilities.

\section{MTMEUR Dataset}
\textcolor{black}{
In this section, we provide a comprehensive description of the MTMEUR dataset, including the data collection, generation, evolution, selection, and human review. Afterwards, we give the detailed dataset statistics.
}

\subsection{Data Collection and Preprocessing}
%
%
We initially collect about 10,000 clips from two websites, pexels\cite{pexels} and mixkit\cite{mixkit}, respectively.
Next, to ensure that the quality and emotion content of the videos meet the research requirements, we implement a rigorous screening process.
Specifically, four reviewers perform an initial evaluation of the video content, aiming to select videos that not only display a single emotion change but also demonstrate emotion interactions within specific contexts. 
Additionally, we consider the diversity of the video backgrounds to ensure that the context provided ample information.
This process also take into account the emotion expressiveness of the video, visual quality, and the effectiveness of background elements.
During the screening process, each reviewer is asked to rate the videos on a scale of 1 to 5, quantifying their evaluation results.
To ensure consistency in the assessments, all reviewers follow clear guidelines when assigning scores and more details can be found in supplementary materials.
After completing the scoring process, we select the top 1,451 highest-rated videos.

\begin{figure}[t]
  \centering
   \includegraphics[width=0.8\linewidth]{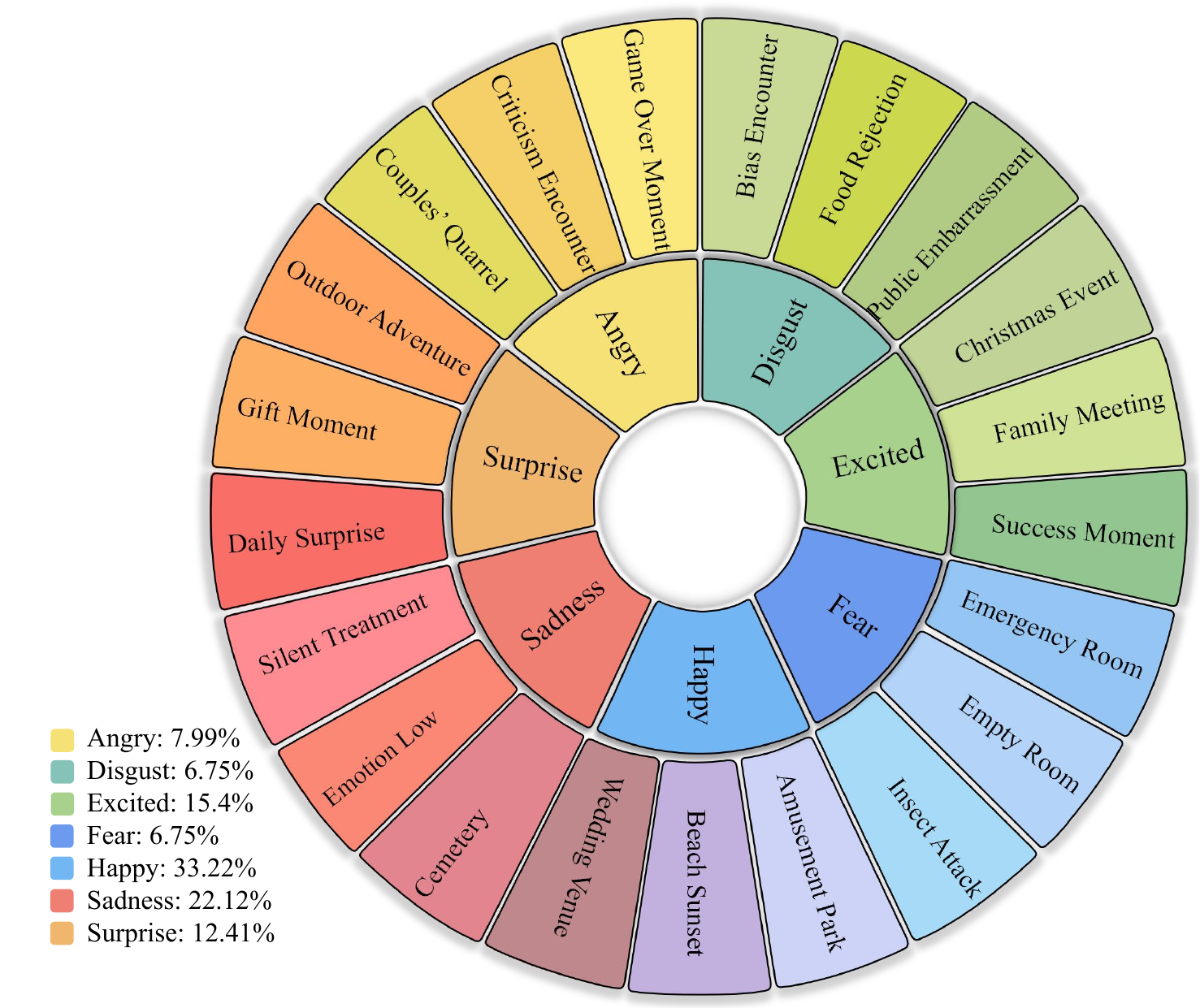}
   \caption{Distribution of video categories. The MTMEUR encompasses videos portraying seven distinct emotions, with each emotional category represented across a diverse range of scenes.}
   \label{fig:emotion_type}
   \vspace{-1em}
\end{figure}


\subsection{Data Generation}
\label{sec:intro}

\begin{figure*}[t]
\centering
\includegraphics[width=0.95\textwidth]{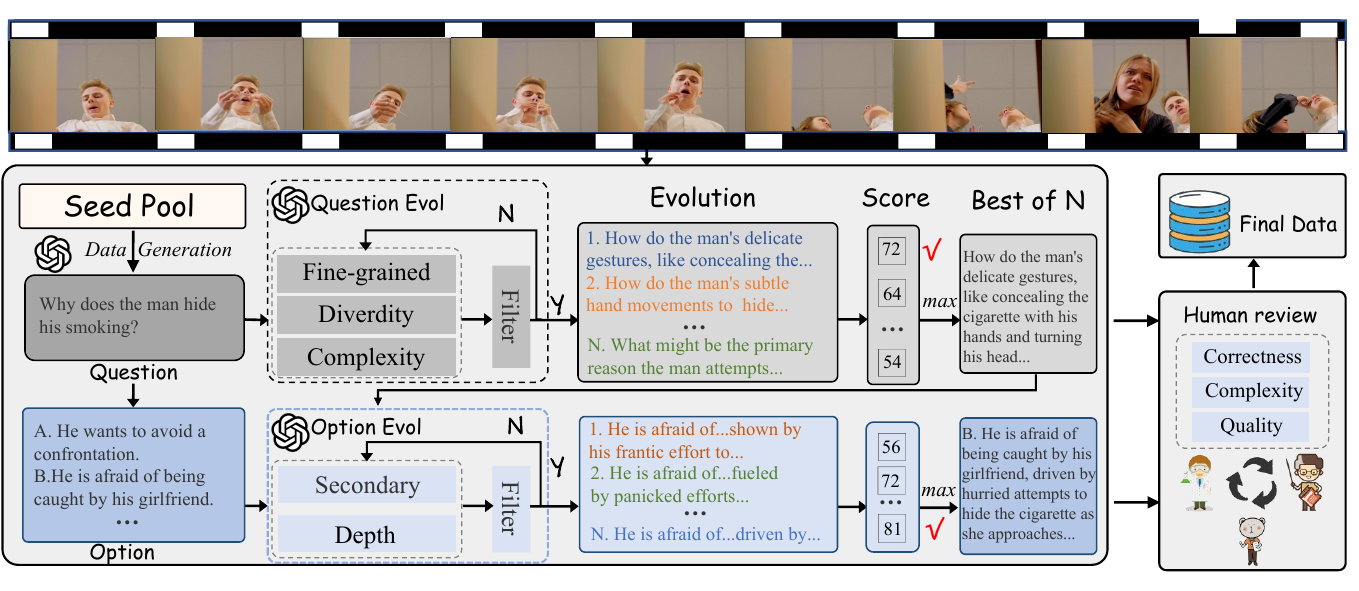}
\caption{
Data Generation Pipeline. The pipeline for data generation comprises, which includes initial creation, iterative evolution, data selection, and manual evaluation.
}
\label{fig:qa_generation}
\end{figure*}

Generating large-scale sentiment-annotated data manually is challenging due to its time-consuming and labor-intensive nature, demanding significant manual effort and meticulous attention to detail.
To tackle this challenge, we developed a unified and automated process to generate high-quality data associated with the videos.


%
In the first step, we select a subset of pre-filtered seed tasks from Social-IQ \cite{zadeh2019social} and further supplement them with 75 manually constructed examples.
These high-quality examples cover diverse emotional reasoning types and serve as few-shot examples to guide GPT-4o to produce better question-answer pairs.
For each video, we randomly sample six tasks from the seed pool and input both video and chosen tasks to GPT-4o to generate two more complex questions.
This method allows us to efficiently generate progressive questions for each video during the initial phase.
These questions cover not only common topics, ranging from fundamental areas like emotion recognition and the triggers of emotion responses, to open-ended inquiries that require reasoning.
For instance, some questions explore the potential outcomes of a situation or ask for suggestions based on specific contexts. 
We generate both the correct option and several distracting options for each question, ensuring that each question has four distinct choices.
These distractor options are carefully designed to resemble the correct answers in surface form while differing conceptually or even being opposites, thereby enhancing their ability to mislead.

\subsection{Question and Option Evolution}

In the first phase of data generation, the generated questions and answers may be relatively simple for MLLMs to understand and answer, and the style of the data remains quite similar to the initial seed data. Motivated by instruction evolution that enhanced instructions can empower LLM better alignment with human preference~\cite{xu2023wizardlmempoweringlargelanguage}, we introduce an iterative refinement mechanism aimed at improving the complexity and diversity of the initial question set.
By utilizing specially designed prompts, we gradually increase the complexity and diversity of the generated questions, making them more aligned with real-world emotion reasoning scenarios. 
Specifically, our iterative mechanism enhances the questions and answers from multiple dimensions, progressively elevating their sophistication to better simulate the complex challenges found in real emotion reasoning tasks.
Additionally, during evolution, a filtering mechanism evaluates the evolved data, re-evolving any that fails to meet standards.

\paragraph{Question Evolution}

We evolve the questions in the following aspects.
%
%
For complexity, we introduce multiple reasoning steps and add constraints based on the specific contexts within the video scenes.
These constraints not only require the model to identify basic emotions but also to understand the complex relationships and causal links between emotions and their contexts.
For example, while an initial question might simply ask about a single emotion, after several iterations, the question may require the model to handle the interplay of multiple emotions during the reasoning process. 
For diversity, we design prompts that guide the model to create coherent and diverse questions across various contexts.
Through multi-turn interactions, these questions explore how emotions evolve at different stages and their impact on other characters within the scenes.
In addition to complexity and diversity, we also focus on fine-grained emotion reasoning, emphasizing the subtle dynamics of emotion shifts.
The emotions depicted in videos are often complex and multi-dimensional, with multiple emotions sometimes overlapping.
To address this, our prompts are designed to guide the model in capturing these subtle emotion changes, ensuring that the generated questions delve deeply into the interaction between emotions.

\paragraph{Answer Evolution}
In addition to question evolution, we further optimize the answers to encompass more contextual information and enhance the reasoning requirements of the options from two aspects.
To increase the depth of the options, we first introduce more constraints by adding qualifiers and modifiers to make the description more specific.
%
Besides, the evolution process involve integrating secondary conditions or details that are closely related to the question context, including emotion shifts in the character, subtle behaviors, and video details.
For example, the original option might be ``The character feels angry due to a friend's misunderstanding.'' 
With increased reasoning depth, the option could be revised to ``The character feels angry due to a friend's misunderstanding but displays coldness and distrust when the friend attempts to explain.''
These steps refine the language to better match the context of the video, necessitating a deeper understanding from the model to accurately differentiate between the correct and distractor options.


\subsection{Data Selection}
\textcolor{black}{
After the multi-turn data generation, we further screen the generated emotion question-answer for each video.
We employ a dual scoring mechanism to evaluate and select the data.
In detail, we instruct ChatGPT to assign a complexity score and a quality score to each generated questions and answers.
A higher complexity score indicates responses that require multi-level reasoning, involve intricate emotion relationships in the video, and capture causal links between emotions. 
A higher quality score reflects responses that accurately convey emotion cues while maintaining coherence and clarity in expression.
We multiplied the complexity and quality scores to obtain a composite score for each question.
We ranked all questions for the same video based on their composite scores and selected the highest-scoring questions as the final retained questions for that video.
Afterwards, we combined the corresponding options and applied a similar scoring process to obtain composite scores, ultimately selecting the highest-scoring options as the final ones for each chosen question.
}

\subsection{Quality Control}
\textcolor{black}{
To ensure that the generated data performs well in terms of correctness, complexity, relevance, and practical application, we also incorporate a strict manual quality control process.
The volunteers with relevant domain expertise carefully screened and evaluated the generated questions and corresponding answers.
Reviewers assess the question-answer pair based on the following three dimensions to ensure they meet the expected standards: 
}


\begin{itemize}[leftmargin=*]
    \item Correctness: The question should be factually accurate and unambiguous, answerable based on the video content. Reviewers verify that the correct answer is precise and that distractors are plausible yet clearly incorrect.
    \item Complexity: The question needs to demonstrate the expected depth and difficulty in reasoning.
    \item Quality: The question should closely relate to the video's content and emotion cues, presenting clear and concise language. It should also display a broad range of emotion types and reasoning patterns, encompassing the emotion changes of different characters in the video and their interactions. 
\end{itemize}

\begin{table}[t]
\footnotesize
\centering
\resizebox{.44\textwidth}{!}{
\begin{tabular}{l|l|c}
\toprule[1pt]

%
{\textsc{\textbf{Data}}} & \textsc{\textbf{Type}} & \textsc{\textbf{Length}} 
\\

\multirow{4}{*} {\textsc{Social-IQ2} \cite{siq2}}
& \textsc{question \#} & {7,098} \\
& \textsc{videos \#} & {1,000} \\
& \textsc{Avg. que} & {11.44} \\
& \textsc{Avg. ans} & {10.89} \\
\midrule
\multirow{4}{*} {\textsc{MTMEUR}}
& \textsc{question \#} & {5,101} \\
& \textsc{videos \#} & {1,451} \\
& \textsc{Avg. que} & {19.63} \\
& \textsc{Avg. ans} & {14.27} \\

\bottomrule

\end{tabular}}
\caption{Comparison of the proposed dataset with Social-IQ2, where
\textsc{Avg. que} and \textsc{Avg. wi}) represent the averaged word-based length of question and answer, respectively.
}
\label{Tab:data_statistics}
\vspace{-1em}
\end{table}

We filter out any data that failed to meet the first criterion (i.e., correctness).
Besides, each question is independently evaluated by at least two reviewers in terms of complexity and quality, with each dimension scored on a scale of 1 to 5.
The final score for each question is the average of the two reviewers' scores.
If the score difference between the two reviewers exceed a certain threshold (i.e., more than 2 points), the question is sent to a third reviewer for arbitration.
After scoring, the questions are ranked based on their total scores, and only those exceeding a minimum threshold (i.e., 3) are retained in the dataset.
To ensure the reliability of the manual review process, we assess the level of consistency among these reviewers.
Before the formal manual review, we randomly selected 100 examples and calculate Fleiss' Kappa between reviewers, which yielded a value exceeding 0.81.
This high Kappa score indicates strong agreement among reviewers, demonstrating the effectiveness and reliability of the manual review process.

To further validate the question solvability, we conducted an additional human evaluation with new volunteers who had not participated in the original annotation process. The volunteers achieved an average accuracy of 96.22\% on dataset, demonstrating that questions are solvable with the corresponding videos.
\subsection{Dataset Statistics}
As shown in Figure \ref{fig:emotion_type}, this dataset covers seven emotions: angry, excited, fear, happy, sad, disgust, and surprise, which also displays the distribution proportions for these emotions.
Each emotion category includes a variety of scenes, such as family meetings, dinners, amusement parks, graduation ceremonies, award events and so on, ensuring the diversity and representativeness of the dataset.
Besides, in Table \ref{Tab:data_statistics}, we provide a comparative analysis of MTMEUR with existing benchmarks, such as Social-IQ2.
Each video in MTMEUR dataset includes an average of four questions, with an average question length of 19.63 words.
The average word length of answers in MTMEUR is 14.27, which is longer than Social-IQ2. 
This suggests that options in MTMEUR are generally more detailed and offer greater specificity.
We also find that the average length of correct answers is similar to that of incorrect answers, which helps to reduce bias that might arise from differences in answer length.

\section{Method}
\begin{figure}[!t]
  \centering
   \includegraphics[width=0.99\linewidth]{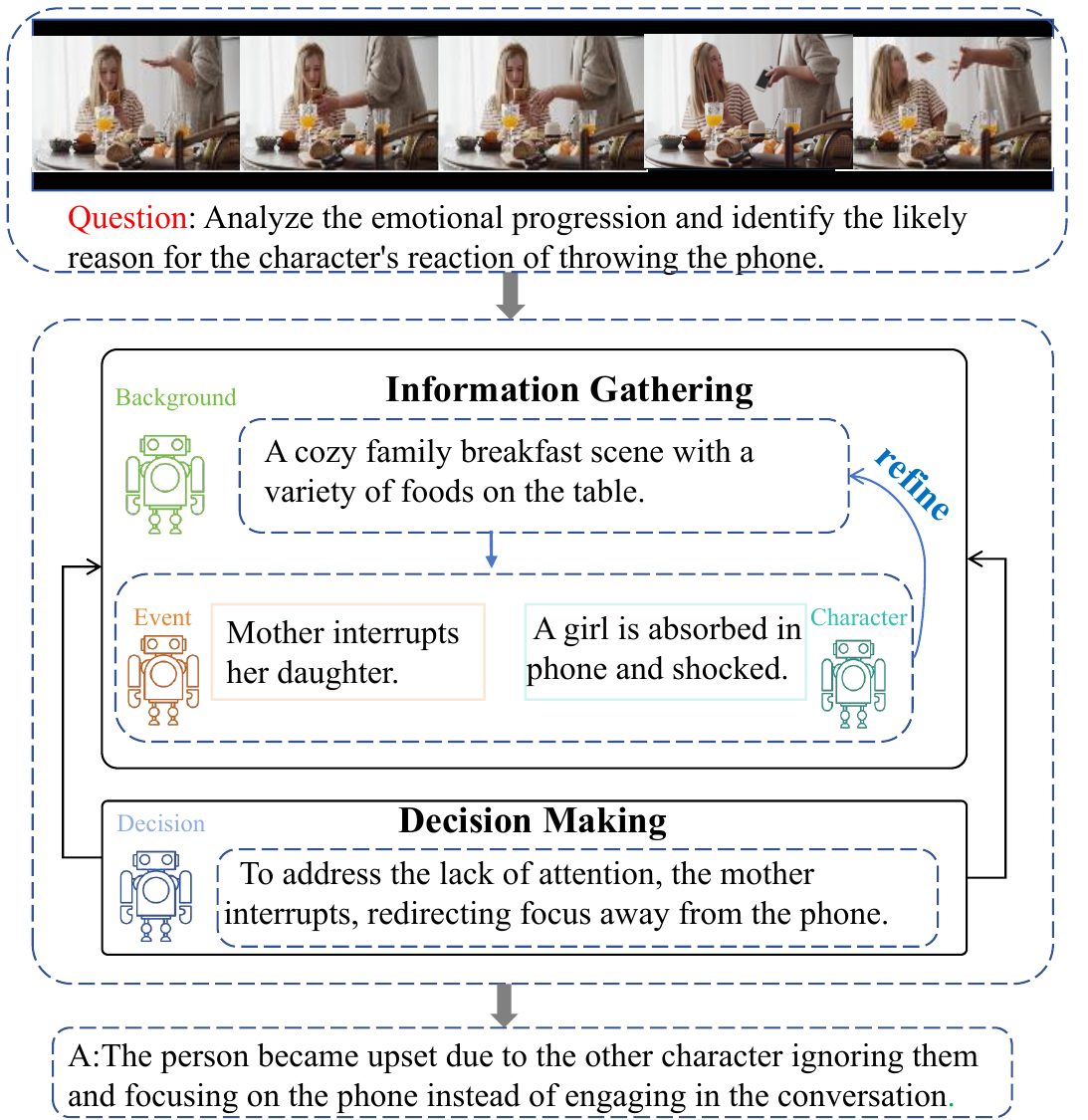}
   \caption{The overall architecture of our method with four agents.}
   \label{fig:model_structure}
   \vspace{-1em}
\end{figure}
Given a video $V$ and $n$ questions $Q$ in natural language along with options $O$, the goal is to select the correct options.
Note that each $O=\{o_{1}, o_{2}, \dots, o_{m}\}$ have $m$ options.
In this section, we present a multi-agent-based multimodal collaboration framework for emotion recognition and reasoning, which consists of four core agents: the background agent, the character agent, the event agent, and the decision agent.  
The former three agents aims to better learn the distinct information from videos through dynamic interaction and collaborative reasoning and the last one is to give the final answer.
An overview of our proposed method is presented in Figure \ref{fig:model_structure}.
%
%

\begin{table*}[ht]
  \centering
  \resizebox{1\textwidth}{!}{
  \begin{tabular}{lccccccc|c}
    \toprule
    \textbf{Model} & \textbf{Disgust} & \textbf{Sadness} & \textbf{Happy} & \textbf{Surprise} & \textbf{Excited} & \textbf{Angry} & \textbf{Fear} & \textbf{Overall ACC} \\ 
    \midrule
    Videochatgpt \cite{maaz2023video} & 27.11 & 24.30 & 28.20 & 23.80 & 36.20 & 26.60 & 27.81 & 29.10 \\ 
    ShareGPT4Video \cite{chen2024sharegpt4video} & 27.24 & 25.60 & 27.90 & 24.30 & 36.10 & 27.20 & 27.68 & 29.39 \\ 
    Chat-UniVi \cite{jin2024chat}  & 29.84 & 24.50 & 30.08 & 25.70 & 35.70 & 24.80 & 29.59 & 30.09 \\ 
    VTimeLLM \cite{huang2024vtimellm} & 33.51 & 21.30 & 34.40 & 34.00 & 46.10 & 28.20 & 33.81 & 34.26 \\ 
    Videollava \cite{lin2023video} & 38.02 & 32.30 & 39.20 & 31.70 & 45.90 & 29.90 & 38.52 & 38.72 \\ 
    Emotion-LLaMA \cite{cheng2024emotion} & 45.52 & 47.10 & 46.60 & 44.30 & 53.50 & 35.40 & 46.13 & 48.65 \\ 
    Qwen-VL-chat \cite{bai2023qwen} & 52.40 & 46.50 & 56.70 & 54.40 & 64.10 & 44.70 & 53.29 & 56.45 \\ 
    MiniCPM-V-2.6 \cite{hu2024minicpm} & 59.15 & 50.90 & 62.75 & 63.80 & 70.80 & 53.60 & 59.42 & 63.21 \\ 
    Videollama2 \cite{cheng2024videollama} & 59.61 & 58.20 & 65.75 & 64.40 & 72.40 & 53.32 & 59.26 & 66.14 \\ 
    Qwen2-VL \cite{wang2024qwen2} & 66.97 & 59.67 & 68.84 & 67.00 & 85.30 & 56.34 & 68.46 & 71.19 \\ 
    Our Method & 68.32 & 60.08 & 71.90 & 69.20 & 87.20 & 56.50 & 69.31 & 72.93 \\
    \bottomrule
  \end{tabular}
  }
  \caption{Comparison of emotion-specific accuracy across different models.}
  \label{tab:emotion-accuracy-comparison-updated}
  \vspace{-1em}
\end{table*}

\subsection{Information Gathering}
Background information is usually important for emotion understanding and reasoning.
For example, in most situations, people in a park are usually happy while in a hospital tend to sad.
Therefore, we first construct a background agent, which focuses on extracting and structuring environmental information from the video.
The background agent employs a tailored prompt to guide the MLLM to produce detailed, structured descriptions of the scene, covering elements such as scene type, object layout, and additional contextual details, denoted as $B_{t}$ at $t$-turn communication.
Next, according to background information, we build the character agent to identifies characters and gives their emotion and psychological states from videos by analyzing facial expressions and body posture.
\begin{equation}
    C_{t} = ChaAgent (V, P_c, q_{i}, o_{i}, H_{t-1})
\end{equation}
where $C$ is the description of character dynamics, $P_c$ is the prompt for character agent and $H_{t-1}=[B_{t-1}, C_{t-1}, E_{t-1}]$ is the history information from previous agent communication.
Meanwhile, the third agent, event agent is proposed to establish causal relationships between the character's emotional states and the video event sequence.
\begin{equation}
    E_{t} = EveAgent (V, P_e, q_{i}, o_{i}, H_{t-1})
\end{equation}
where $E$ is the description of event details, $P_e$ is the prompt for event agent.
In doing so, the three agents engage in collaborative interactions to refine the quality of their individual descriptions.
Sub
For example, character agent take the event into consideration and provides a specific and comprehensive character description.
\subsection{Decision Agent}
We gather diverse information from the three agents, each specializing in different aspects of the video.
The decision agent functions as the ultimate decision-making unit within the system, tasked with synthesizing background information, character dynamics, and event details to formulate the final answer to the questions.
During the decision-making process, the decision agent combines contextual information $H = [H_{1}, H_{2}, \cdots, H_{T}]$ and give the correct option, where $T$ is the number of interaction turns within the multi-agent framework.

\section{Experiments}
%
\subsection{Baselines and Evaluation Metrics}
To explore the performance of different models, we evaluate a range of representative open-source multimodal large models, including Video-ChatGPT \cite{maaz2023video}, Video-LLaVA \cite{lin2023video}, MiniCPM \cite{hu2024minicpm}, VTimeLLM \cite{huang2024vtimellm}, Emotion-LLaMA \cite{cheng2024emotion}, ShareGPT4Video \cite{chen2024sharegpt4video}, VideoLLaMA2 \cite{cheng2024videollama}, Chat-UniVi \cite{jin2024chat}, Qwen-VL-chat \cite{bai2023qwen}, and Qwen2-VL \cite{wang2024qwen2}.

The proposed dataset contain both single-choice and multiple-choice questions.
We use accuracy to evaluate the model performance, which is defined as the ratio of completely correct answers to the total number of questions.
\subsection{Implementation Details}
In our experiments, we use the Autogen\cite{wu2024autogen} library to build our agent-based framework.
Specifically, we adopt Qwen2-VL as the backbone model for our proposed approach.
%
%
%
During inference, we set the temperature to 0 in all experiments to control randomness.
Besides, our experiments are conducted on the A6000 GPUs.
%
%

\begin{table}[t]
 \resizebox{0.45\textwidth}{!}{%
  \centering
  \begin{tabular}{lll|c}
    \toprule
    \multicolumn{3}{c|}{Configuration} & \multirow{2}{*}{Accuracy (\%)} \\
    \cmidrule(l){1-3}
    
    Background & Character & Event & \\

    \midrule
    \checkmark & \checkmark  & & 72.16\\
    \checkmark & & \checkmark & 71.79\\
     & \checkmark & \checkmark & 71.27\\
\checkmark & \checkmark & \checkmark & 72.93\\
    \bottomrule
  \end{tabular}}
  \caption{Ablation study showing the effect of different agents.}
  \label{tab:ablation_new}
  \vspace{-1em}
\end{table}
\subsection{Overall Performance}
To explore the performance of different models in our proposed dataset, we report the results in Table \ref{tab:emotion-accuracy-comparison-updated}.
We show both their overall accuracy and their respective results across different emotion categories. 
There are several observation drawn from the results.
First, Qwen2-VL leads with an overall accuracy of 71.19\%, outperforming other models, demonstrating its strong capabilities in emotion recognition and emotion reasoning. 
Second, our proposed model outperform its base model (i.e., Qwen2-VL), which also performs exceptionally well, achieving an accuracy of 72.93\%.
This illustrates that using different agents to integrate background, character, and event information is beneficial to improve emotion reasoning.
Third, in terms of emotion categories, most positive emotions such as Happy and Surprise tend to be recognized more accurately.
These positive emotions are often accompanied by clear facial expressions, and semantically easy-to-interpret cues, which making it easier for models to extract emotion information.
In contrast, most negative emotions are generally with relatively lower accuracy, especially Angry and Sadness, which consistently shows poorer performance due to its inherent complexity and subtle emotion cues.

\subsection{Ablation Study}
To explore effect of different agent in this task, we conduct ablation study and the results are reported in Table \ref{tab:ablation_new}.
It is observed that removing any agent from our proposed method results in a noticeable decline in performance, highlighting the contribution of each agent in enhancing the model's ability to comprehend the video content.
Notably, the removal of the Background agent leads to a significant accuracy drop from 72.93\% to 71.27\%.
This highlights the importance of background and contextual information, which serves as a foundation for understanding character actions and event development.

\subsection{Effect of Question Type}
\begin{table}[t]
\centering
\resizebox{0.5\textwidth}{!}{%
\begin{tabular}{l|c|c|c|c|c}
    \toprule
    \textbf{Model} & \textbf{Ana.} & \textbf{InDir.} & \textbf{Dir.} & \textbf{Pre.} & \textbf{Oth.} \\ 
    \midrule
    Emotion-LLaMA  & 50.59 & 46.88 & 53.06 & 50.10 & 54.01 \\ 
    Qwen2-VL       & 71.10 & 68.20  & 73.90  & 66.67 & 75.92 \\ 
    Our Method     & 72.73 & 71.50  & 76.00  & 68.20 & 77.67 \\ 
    \bottomrule
\end{tabular}
}
\caption{Comparison of category-specific accuracy across different models, where \textbf{Ana.}, \textbf{InDir.}, \textbf{Dir.}, \textbf{Pre.}, and \textbf{Oth.} refer to Current State, Indirect Causality, Direct Causality, Prediction of Future, and Others, respectively.}
\label{tab:category-accuracy-comparison}
\end{table}

\begin{table}[t]
\centering
\resizebox{0.5\textwidth}{!}{%
\begin{tabular}{l|c|c}
    \toprule
    \textbf{Model} & \textbf{Before Evolution (\%)} & \textbf{After Evolution (\%)}  \\ 
    \midrule
    VTimeLLM        & 39.82 & 34.26 \\  
    MiniCPM-V-2.6   & 71.47 & 63.21 \\ 
    VideoLLaMA2     & 79.09 & 66.14 \\ 
    Qwen2-VL        & 83.47 & 71.19 \\
    \bottomrule
\end{tabular}
}
\caption{Accuracy comparison before and after evolution .}
\label{tab:evolution_comparison}
\vspace{-1em}

\end{table}

\textcolor{black}{
To investigate the impact of different question categories on model performance, we evaluate the performance of different models across five question categories in the MTMEUR dataset: Current State, Direct Causality, Indirect Causality, Prediction of Future and others. The results are reported in Table \ref{tab:category-accuracy-comparison}. 
It is observed that that Prediction of Future and Indirect Causality show the lowest accuracies, reflecting the challenges models face in handling complex reasoning tasks.
These categories require not only recognizing emotions in the current context but also predicting how they evolve over time or how they are influenced by multiple factors. 
In contrast, the Current State and Direct Causality categories yield relatively higher performance, as these tasks primarily involve recognizing and reasoning based on observable emotional states and direct relationships between emotions and events.
}

\begin{table}[t]
\centering
\resizebox{0.38\textwidth}{!}{%
\begin{tabular}{l|c|c}
    \toprule
    \textbf{Prompt Type} & \textbf{Result (\%)} & \textbf{$\Delta$}\\ 
    \midrule
    Qwen2-VL  & 71.19 & - \\ 
    Step-Back \cite{zheng2024take} & 70.57 & -0.62\% \\  
    AP \cite{yasunaga2024large} & 70.68 & -0.51\% \\ 
    Few-shot \cite{brown2020language} & 71.84 & +0.65\% \\ 
    CoT \cite{wei2022chain} & 71.98 & +0.79\% \\ 
    Our Method & 72.93 & +1.74\% \\
    \bottomrule
\end{tabular}
}
\caption{Performance of different prompt methods.}
\label{tab:prompt_results}
\vspace{-2em}
\end{table}

\begin{figure*}[ht]
\centering
\includegraphics[width=0.98\textwidth]{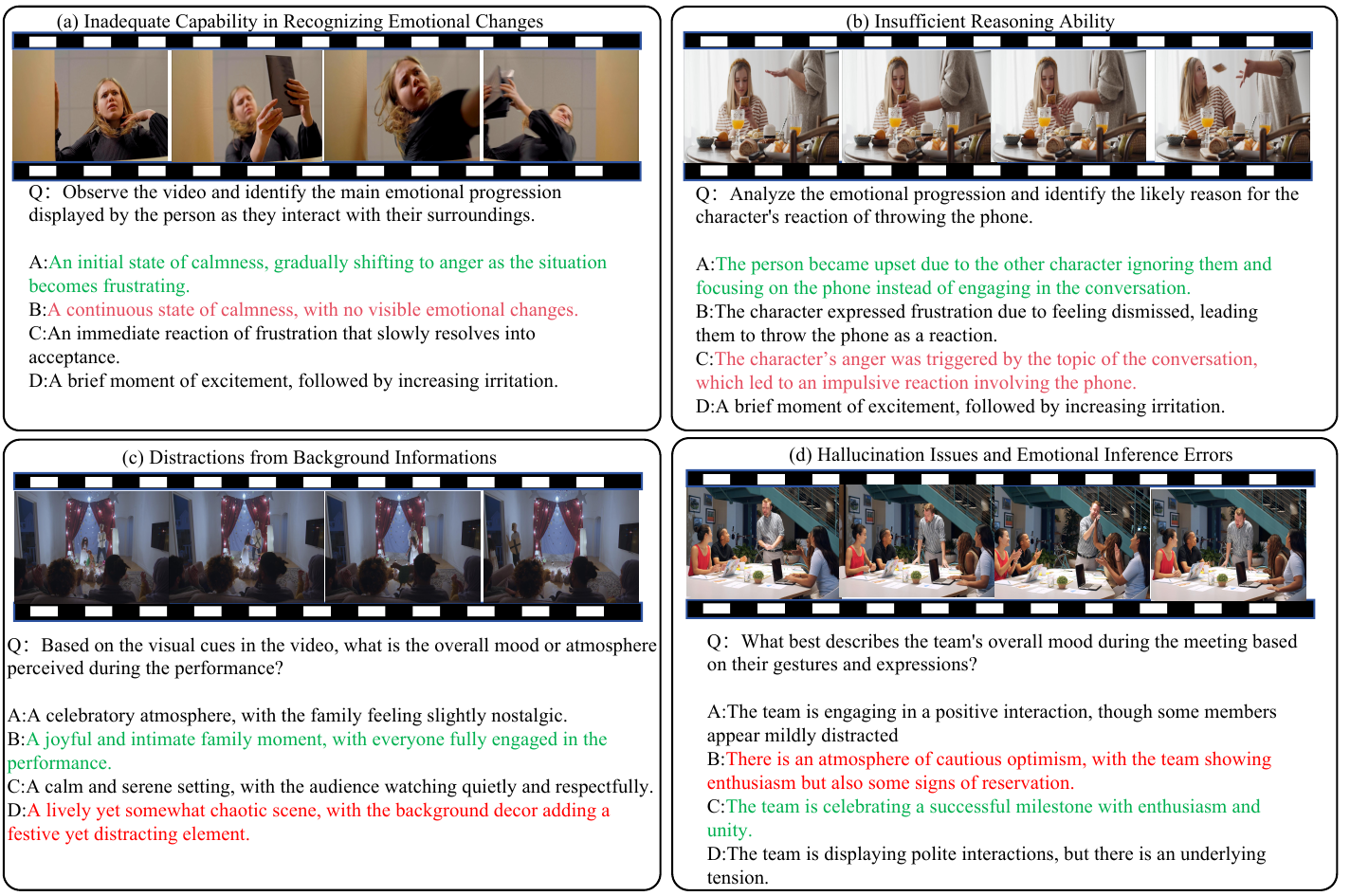}
\caption{Examples of mistakes made by the model on the dataset. Text in red represents the model's choice, text in green represents the correct answer.
}

\label{fig:error_analysis}
\end{figure*}
\subsection{Effect of Evolution}
To explore the effect of the question and option evolution in this task,  we conduct an experiment and report the results in Table \ref{tab:evolution_comparison}. 
We show the performance of the our dataset before and after evolution across different models. 
It is observed that all models experience a noticeable decline in accuracy after the evolution of the dataset. 
This decline can be attributed to the increased complexity and diversity of the evolved dataset, which are designed to challenge the model's emotion reasoning capabilities more effectively. 
\textcolor{black}{
Specifically, the average length of questions increases from 11.21 words to 19.63 words, and the average length of options grows from 10.94 words to 14.27 words. 
The increased question length introduces more contextual information and necessitates multi-step reasoning, while the longer options require models to make finer distinctions between emotional states, collectively raising the difficulty of accurately interpreting emotional cues in multi-turn reasoning scenarios.
}
\subsection{Effect of Different Prompt}
To explore the impact of the different prompts on Qwen2-VL in our proposed dataset, we conduct experiments using four distinct prompting strategies: Few-Shot \cite{brown2020language}, Chain-of-Thought(CoT) \cite{wei2022chain}, Analogical Prompting (AP) \cite{yasunaga2024large}, and Step-Back \cite{zheng2024take}.
The results are shown in Table \ref{tab:prompt_results}.
It is observed that few-shot yield improvements over baseline, which illustrate that appropriate examples are helpful for the model to understand the video and generate correct responses.
Meanwhile, the comparison between CoT and base model shows that enabling the model to follow a reasoning pathways when answering questions is helpful for improving the reasoning ability of the model.
However, AP and Step-Back have a slight decrease compared to base model.
Analogical Prompting relies heavily on the quality and relevance of retrieved analogies, while generating appropriate exemplars is challenging in multimodal emotion reasoning.
Step-Back Prompting, which encourages the model to "step back" before answering, can sometimes lead to deviations from the optimal reasoning path. 
This additional reasoning step may cause the model to overthink, ultimately impacting performance.
%


\subsection{Error Analysis}

Based on our experiments on the MTMEUR dataset, we find several typical issues, as detailed below.
1) The model exhibits significant limitations in recognizing dynamic emotion shifts, especially in the video contexts involving multiple emotion interactions and emotion changes.
As shown in Fig. \ref{fig:error_analysis} (a), where the girl transitions from calmness to anger, the model frequently identifies only the initial calm state, neglecting the critical shift toward anger that follows.
2) While the model occasionally identifies basic emotional states in the video, its performance declines when dealing with complex reasoning tasks.
As shown in the first row of Fig. \ref{fig:error_analysis} (b), the model may detect a character's anger but fails to associate this emotion with subsequent behaviors or events.
3) The responses of the model are often affected by background information in the video, especially in complex contexts or when the visual background contradicts emotion cues. As shown in the second row of Fig. \ref{fig:error_analysis} (c), the model might mistakenly select "a lively and chaotic scene" due to the bright, festive decorations in the video background.
4) The model sometimes generates “hallucinations,” producing nonexistent emotions or scenes. In the second row of Fig. \ref{fig:error_analysis} (d), the model relies on certain body language of the team members, mistakenly infers complex emotional dynamics, and fails to recognize that the main atmosphere of the scene is celebratory.

\section{Discussion and Conclusion} 
In this paper, we introduce MTMEUR, a rigorously designed multimodal emotion understanding and reasoning benchmark.
The dataset consists of 5,101 multiple-choice questions tailored to evaluate the emotion reasoning proficiency of MLLMs within a broad range of intricate real-world video scenarios.
Moreover, we propose a multi-agent-based multimodal collaboration framework, where different agents focus on different aspects.
These agents interact and refine each other's descriptions to improve information gathering so as to enhance reasoning ability.
We further present an experimental evaluation of several popular MLLMs using the MTMEUR benchmark, which reveals that most existing models continue to face significant challenges in this task.
Future work can leverage MTMEUR to further refine MLLMs' performance in these complex scenarios, improving their application in real-world emotion understanding and reasoning tasks.
An intriguing finding from our experiments is that reasoning about negative emotions tends to be more challenging than reasoning about positive emotions. 
This observation highlights a key challenge in applying MLLMs. 

\section*{Acknowledgments}

This work was supported in part by National Natural Science Foundation of China under Grant 62402158, Grant 72188101 and Grant U22A2094; by the Major Project of Anhui Province Grant 202203a05020011, Grant 202423k09020001, and Grant 2408085J040; and by the Fundamental Research Funds for the Central Universities Grant JZ2024HGTG0309, Grant JZ2024AHST0337, Grant JZ2023YQ-TD0072.

\bibliographystyle{ACM-Reference-Format}
\bibliography{sample-base}


\begin{thebibliography}{64}


\ifx \showCODEN    \undefined \def \showCODEN     #1{\unskip}     \fi
\ifx \showDOI      \undefined \def \showDOI       #1{#1}\fi
\ifx \showISBNx    \undefined \def \showISBNx     #1{\unskip}     \fi
\ifx \showISBNxiii \undefined \def \showISBNxiii  #1{\unskip}     \fi
\ifx \showISSN     \undefined \def \showISSN      #1{\unskip}     \fi
\ifx \showLCCN     \undefined \def \showLCCN      #1{\unskip}     \fi
\ifx \shownote     \undefined \def \shownote      #1{#1}          \fi
\ifx \showarticletitle \undefined \def \showarticletitle #1{#1}   \fi
\ifx \showURL      \undefined \def \showURL       {\relax}        \fi
\providecommand\bibfield[2]{#2}
\providecommand\bibinfo[2]{#2}
\providecommand\natexlab[1]{#1}
\providecommand\showeprint[2][]{arXiv:#2}

\bibitem[Abdullah et~al\mbox{.}(2021)]%
        {abdullah2021multimodal}
\bibfield{author}{\bibinfo{person}{Sharmeen M Saleem~Abdullah Abdullah}, \bibinfo{person}{Siddeeq Y~Ameen Ameen}, \bibinfo{person}{Mohammed~AM Sadeeq}, {and} \bibinfo{person}{Subhi Zeebaree}.} \bibinfo{year}{2021}\natexlab{}.
\newblock \showarticletitle{Multimodal emotion recognition using deep learning}.
\newblock \bibinfo{journal}{\emph{Journal of Applied Science and Technology Trends}} \bibinfo{volume}{2}, \bibinfo{number}{01} (\bibinfo{year}{2021}), \bibinfo{pages}{73--79}.
\newblock


\bibitem[Bai et~al\mbox{.}(2023)]%
        {bai2023qwen}
\bibfield{author}{\bibinfo{person}{Jinze Bai}, \bibinfo{person}{Shuai Bai}, \bibinfo{person}{Shusheng Yang}, \bibinfo{person}{Shijie Wang}, \bibinfo{person}{Sinan Tan}, \bibinfo{person}{Peng Wang}, \bibinfo{person}{Junyang Lin}, \bibinfo{person}{Chang Zhou}, {and} \bibinfo{person}{Jingren Zhou}.} \bibinfo{year}{2023}\natexlab{}.
\newblock \showarticletitle{Qwen-vl: A frontier large vision-language model with versatile abilities}.
\newblock \bibinfo{journal}{\emph{arXiv preprint arXiv:2308.12966}} (\bibinfo{year}{2023}).
\newblock


\bibitem[Brown et~al\mbox{.}(2020)]%
        {brown2020language}
\bibfield{author}{\bibinfo{person}{Tom Brown}, \bibinfo{person}{Benjamin Mann}, \bibinfo{person}{Nick Ryder}, \bibinfo{person}{Melanie Subbiah}, \bibinfo{person}{Jared~D Kaplan}, \bibinfo{person}{Prafulla Dhariwal}, \bibinfo{person}{Arvind Neelakantan}, \bibinfo{person}{Pranav Shyam}, \bibinfo{person}{Girish Sastry}, \bibinfo{person}{Amanda Askell}, {et~al\mbox{.}}} \bibinfo{year}{2020}\natexlab{}.
\newblock \showarticletitle{Language models are few-shot learners}.
\newblock \bibinfo{journal}{\emph{Advances in neural information processing systems}}  \bibinfo{volume}{33} (\bibinfo{year}{2020}), \bibinfo{pages}{1877--1901}.
\newblock


\bibitem[Busso et~al\mbox{.}(2008)]%
        {busso2008iemocap}
\bibfield{author}{\bibinfo{person}{Carlos Busso}, \bibinfo{person}{Murtaza Bulut}, \bibinfo{person}{Chi-Chun Lee}, \bibinfo{person}{Abe Kazemzadeh}, \bibinfo{person}{Emily Mower}, \bibinfo{person}{Samuel Kim}, \bibinfo{person}{Jeannette~N Chang}, \bibinfo{person}{Sungbok Lee}, {and} \bibinfo{person}{Shrikanth~S Narayanan}.} \bibinfo{year}{2008}\natexlab{}.
\newblock \showarticletitle{IEMOCAP: Interactive emotional dyadic motion capture database}.
\newblock \bibinfo{journal}{\emph{Language resources and evaluation}}  \bibinfo{volume}{42} (\bibinfo{year}{2008}), \bibinfo{pages}{335--359}.
\newblock


\bibitem[Chen et~al\mbox{.}(2024)]%
        {chen2024sharegpt4video}
\bibfield{author}{\bibinfo{person}{Lin Chen}, \bibinfo{person}{Xilin Wei}, \bibinfo{person}{Jinsong Li}, \bibinfo{person}{Xiaoyi Dong}, \bibinfo{person}{Pan Zhang}, \bibinfo{person}{Yuhang Zang}, \bibinfo{person}{Zehui Chen}, \bibinfo{person}{Haodong Duan}, \bibinfo{person}{Bin Lin}, \bibinfo{person}{Zhenyu Tang}, {et~al\mbox{.}}} \bibinfo{year}{2024}\natexlab{}.
\newblock \showarticletitle{Sharegpt4video: Improving video understanding and generation with better captions}.
\newblock \bibinfo{journal}{\emph{arXiv preprint arXiv:2406.04325}} (\bibinfo{year}{2024}).
\newblock


\bibitem[Cheng et~al\mbox{.}(2024a)]%
        {cheng2024emotion}
\bibfield{author}{\bibinfo{person}{Zebang Cheng}, \bibinfo{person}{Zhi-Qi Cheng}, \bibinfo{person}{Jun-Yan He}, \bibinfo{person}{Jingdong Sun}, \bibinfo{person}{Kai Wang}, \bibinfo{person}{Yuxiang Lin}, \bibinfo{person}{Zheng Lian}, \bibinfo{person}{Xiaojiang Peng}, {and} \bibinfo{person}{Alexander Hauptmann}.} \bibinfo{year}{2024}\natexlab{a}.
\newblock \showarticletitle{Emotion-LLaMA: Multimodal Emotion Recognition and Reasoning with Instruction Tuning}.
\newblock \bibinfo{journal}{\emph{arXiv preprint arXiv:2406.11161}} (\bibinfo{year}{2024}).
\newblock


\bibitem[Cheng et~al\mbox{.}(2024b)]%
        {cheng2024videollama}
\bibfield{author}{\bibinfo{person}{Zesen Cheng}, \bibinfo{person}{Sicong Leng}, \bibinfo{person}{Hang Zhang}, \bibinfo{person}{Yifei Xin}, \bibinfo{person}{Xin Li}, \bibinfo{person}{Guanzheng Chen}, \bibinfo{person}{Yongxin Zhu}, \bibinfo{person}{Wenqi Zhang}, \bibinfo{person}{Ziyang Luo}, \bibinfo{person}{Deli Zhao}, {et~al\mbox{.}}} \bibinfo{year}{2024}\natexlab{b}.
\newblock \showarticletitle{VideoLLaMA 2: Advancing Spatial-Temporal Modeling and Audio Understanding in Video-LLMs}.
\newblock \bibinfo{journal}{\emph{arXiv preprint arXiv:2406.07476}} (\bibinfo{year}{2024}).
\newblock


\bibitem[Chowdhery et~al\mbox{.}(2023)]%
        {chowdhery2023palm}
\bibfield{author}{\bibinfo{person}{Aakanksha Chowdhery}, \bibinfo{person}{Sharan Narang}, \bibinfo{person}{Jacob Devlin}, \bibinfo{person}{Maarten Bosma}, \bibinfo{person}{Gaurav Mishra}, \bibinfo{person}{Adam Roberts}, \bibinfo{person}{Paul Barham}, \bibinfo{person}{Hyung~Won Chung}, \bibinfo{person}{Charles Sutton}, \bibinfo{person}{Sebastian Gehrmann}, {et~al\mbox{.}}} \bibinfo{year}{2023}\natexlab{}.
\newblock \showarticletitle{Palm: Scaling language modeling with pathways}.
\newblock \bibinfo{journal}{\emph{Journal of Machine Learning Research}} \bibinfo{volume}{24}, \bibinfo{number}{240} (\bibinfo{year}{2023}), \bibinfo{pages}{1--113}.
\newblock


\bibitem[Coppersmith et~al\mbox{.}(2015)]%
        {coppersmith2015clpsych}
\bibfield{author}{\bibinfo{person}{Glen Coppersmith}, \bibinfo{person}{Mark Dredze}, \bibinfo{person}{Craig Harman}, \bibinfo{person}{Kristy Hollingshead}, {and} \bibinfo{person}{Margaret Mitchell}.} \bibinfo{year}{2015}\natexlab{}.
\newblock \showarticletitle{CLPsych 2015 shared task: Depression and PTSD on Twitter}. In \bibinfo{booktitle}{\emph{Proceedings of the 2nd workshop on computational linguistics and clinical psychology: from linguistic signal to clinical reality}}. \bibinfo{pages}{31--39}.
\newblock


\bibitem[Fang et~al\mbox{.}(2022)]%
        {fang2022faf}
\bibfield{author}{\bibinfo{person}{Zhongyu Fang}, \bibinfo{person}{Aoyun He}, \bibinfo{person}{Qihui Yu}, \bibinfo{person}{Baopeng Gao}, \bibinfo{person}{Weiping Ding}, \bibinfo{person}{Tong Zhang}, {and} \bibinfo{person}{Lei Ma}.} \bibinfo{year}{2022}\natexlab{}.
\newblock \showarticletitle{FAF: A novel multimodal emotion recognition approach integrating face, body and text}.
\newblock \bibinfo{journal}{\emph{arXiv preprint arXiv:2211.15425}} (\bibinfo{year}{2022}).
\newblock


\bibitem[Goel et~al\mbox{.}(2023)]%
        {goel2023llms}
\bibfield{author}{\bibinfo{person}{Akshay Goel}, \bibinfo{person}{Almog Gueta}, \bibinfo{person}{Omry Gilon}, \bibinfo{person}{Chang Liu}, \bibinfo{person}{Sofia Erell}, \bibinfo{person}{Lan~Huong Nguyen}, \bibinfo{person}{Xiaohong Hao}, \bibinfo{person}{Bolous Jaber}, \bibinfo{person}{Shashir Reddy}, \bibinfo{person}{Rupesh Kartha}, {et~al\mbox{.}}} \bibinfo{year}{2023}\natexlab{}.
\newblock \showarticletitle{Llms accelerate annotation for medical information extraction}. In \bibinfo{booktitle}{\emph{machine learning for health (ML4H)}}. PMLR, \bibinfo{pages}{82--100}.
\newblock


\bibitem[Gratch et~al\mbox{.}(2014)]%
        {gratch2014distress}
\bibfield{author}{\bibinfo{person}{Jonathan Gratch}, \bibinfo{person}{Ron Artstein}, \bibinfo{person}{Gale~M Lucas}, \bibinfo{person}{Giota Stratou}, \bibinfo{person}{Stefan Scherer}, \bibinfo{person}{Angela Nazarian}, \bibinfo{person}{Rachel Wood}, \bibinfo{person}{Jill Boberg}, \bibinfo{person}{David DeVault}, \bibinfo{person}{Stacy Marsella}, {et~al\mbox{.}}} \bibinfo{year}{2014}\natexlab{}.
\newblock \showarticletitle{The distress analysis interview corpus of human and computer interviews.}. In \bibinfo{booktitle}{\emph{LREC}}. Reykjavik, \bibinfo{pages}{3123--3128}.
\newblock


\bibitem[Hu et~al\mbox{.}(2024a)]%
        {hu2024psycollm}
\bibfield{author}{\bibinfo{person}{Jinpeng Hu}, \bibinfo{person}{Tengteng Dong}, \bibinfo{person}{Luo Gang}, \bibinfo{person}{Hui Ma}, \bibinfo{person}{Peng Zou}, \bibinfo{person}{Xiao Sun}, \bibinfo{person}{Dan Guo}, \bibinfo{person}{Xun Yang}, {and} \bibinfo{person}{Meng Wang}.} \bibinfo{year}{2024}\natexlab{a}.
\newblock \showarticletitle{Psycollm: Enhancing llm for psychological understanding and evaluation}.
\newblock \bibinfo{journal}{\emph{IEEE Transactions on Computational Social Systems}} (\bibinfo{year}{2024}).
\newblock


\bibitem[Hu et~al\mbox{.}(2021)]%
        {hu2021word}
\bibfield{author}{\bibinfo{person}{Jinpeng Hu}, \bibinfo{person}{Jianling Li}, \bibinfo{person}{Zhihong Chen}, \bibinfo{person}{Yaling Shen}, \bibinfo{person}{Yan Song}, \bibinfo{person}{Xiang Wan}, {and} \bibinfo{person}{Tsung-Hui Chang}.} \bibinfo{year}{2021}\natexlab{}.
\newblock \showarticletitle{Word Graph Guided Summarization for Radiology Findings}. In \bibinfo{booktitle}{\emph{Findings of the Association for Computational Linguistics: ACL-IJCNLP 2021}}. \bibinfo{pages}{4980--4990}.
\newblock


\bibitem[Hu et~al\mbox{.}(2022a)]%
        {hu2022graph}
\bibfield{author}{\bibinfo{person}{Jinpeng Hu}, \bibinfo{person}{Zhuo Li}, \bibinfo{person}{Zhihong Chen}, \bibinfo{person}{Zhen Li}, \bibinfo{person}{Xiang Wan}, {and} \bibinfo{person}{Tsung-Hui Chang}.} \bibinfo{year}{2022}\natexlab{a}.
\newblock \showarticletitle{Graph Enhanced Contrastive Learning for Radiology Findings Summarization}. In \bibinfo{booktitle}{\emph{Proceedings of the 60th Annual Meeting of the Association for Computational Linguistics (Volume 1: Long Papers)}}. \bibinfo{pages}{4677--4688}.
\newblock


\bibitem[Hu et~al\mbox{.}(2022b)]%
        {hu2022hero}
\bibfield{author}{\bibinfo{person}{Jinpeng Hu}, \bibinfo{person}{Yaling Shen}, \bibinfo{person}{Yang Liu}, \bibinfo{person}{Xiang Wan}, {and} \bibinfo{person}{Tsung-Hui Chang}.} \bibinfo{year}{2022}\natexlab{b}.
\newblock \showarticletitle{Hero-Gang Neural Model For Named Entity Recognition}. In \bibinfo{booktitle}{\emph{Proceedings of the 2022 Conference of the North American Chapter of the Association for Computational Linguistics: Human Language Technologies}}. \bibinfo{pages}{1924--1936}.
\newblock


\bibitem[Hu et~al\mbox{.}(2024b)]%
        {hu2024minicpm}
\bibfield{author}{\bibinfo{person}{Shengding Hu}, \bibinfo{person}{Yuge Tu}, \bibinfo{person}{Xu Han}, \bibinfo{person}{Chaoqun He}, \bibinfo{person}{Ganqu Cui}, \bibinfo{person}{Xiang Long}, \bibinfo{person}{Zhi Zheng}, \bibinfo{person}{Yewei Fang}, \bibinfo{person}{Yuxiang Huang}, \bibinfo{person}{Weilin Zhao}, {et~al\mbox{.}}} \bibinfo{year}{2024}\natexlab{b}.
\newblock \showarticletitle{Minicpm: Unveiling the potential of small language models with scalable training strategies}.
\newblock \bibinfo{journal}{\emph{arXiv preprint arXiv:2404.06395}} (\bibinfo{year}{2024}).
\newblock


\bibitem[Huang et~al\mbox{.}(2024)]%
        {huang2024vtimellm}
\bibfield{author}{\bibinfo{person}{Bin Huang}, \bibinfo{person}{Xin Wang}, \bibinfo{person}{Hong Chen}, \bibinfo{person}{Zihan Song}, {and} \bibinfo{person}{Wenwu Zhu}.} \bibinfo{year}{2024}\natexlab{}.
\newblock \showarticletitle{Vtimellm: Empower llm to grasp video moments}. In \bibinfo{booktitle}{\emph{Proceedings of the IEEE/CVF Conference on Computer Vision and Pattern Recognition}}. \bibinfo{pages}{14271--14280}.
\newblock


\bibitem[Jin et~al\mbox{.}(2024)]%
        {jin2024chat}
\bibfield{author}{\bibinfo{person}{Peng Jin}, \bibinfo{person}{Ryuichi Takanobu}, \bibinfo{person}{Wancai Zhang}, \bibinfo{person}{Xiaochun Cao}, {and} \bibinfo{person}{Li Yuan}.} \bibinfo{year}{2024}\natexlab{}.
\newblock \showarticletitle{Chat-univi: Unified visual representation empowers large language models with image and video understanding}. In \bibinfo{booktitle}{\emph{Proceedings of the IEEE/CVF Conference on Computer Vision and Pattern Recognition}}. \bibinfo{pages}{13700--13710}.
\newblock


\bibitem[Li et~al\mbox{.}(2024b)]%
        {li2024llava}
\bibfield{author}{\bibinfo{person}{Chunyuan Li}, \bibinfo{person}{Cliff Wong}, \bibinfo{person}{Sheng Zhang}, \bibinfo{person}{Naoto Usuyama}, \bibinfo{person}{Haotian Liu}, \bibinfo{person}{Jianwei Yang}, \bibinfo{person}{Tristan Naumann}, \bibinfo{person}{Hoifung Poon}, {and} \bibinfo{person}{Jianfeng Gao}.} \bibinfo{year}{2024}\natexlab{b}.
\newblock \showarticletitle{Llava-med: Training a large language-and-vision assistant for biomedicine in one day}.
\newblock \bibinfo{journal}{\emph{Advances in Neural Information Processing Systems}}  \bibinfo{volume}{36} (\bibinfo{year}{2024}).
\newblock


\bibitem[Li et~al\mbox{.}(2024c)]%
        {li2024llava2}
\bibfield{author}{\bibinfo{person}{Feng Li}, \bibinfo{person}{Renrui Zhang}, \bibinfo{person}{Hao Zhang}, \bibinfo{person}{Yuanhan Zhang}, \bibinfo{person}{Bo Li}, \bibinfo{person}{Wei Li}, \bibinfo{person}{Zejun Ma}, {and} \bibinfo{person}{Chunyuan Li}.} \bibinfo{year}{2024}\natexlab{c}.
\newblock \showarticletitle{LLaVA-NeXT-Interleave: Tackling Multi-image, Video, and 3D in Large Multimodal Models}.
\newblock \bibinfo{journal}{\emph{arXiv preprint arXiv:2407.07895}} (\bibinfo{year}{2024}).
\newblock


\bibitem[Li et~al\mbox{.}(2023)]%
        {li2023videochat}
\bibfield{author}{\bibinfo{person}{KunChang Li}, \bibinfo{person}{Yinan He}, \bibinfo{person}{Yi Wang}, \bibinfo{person}{Yizhuo Li}, \bibinfo{person}{Wenhai Wang}, \bibinfo{person}{Ping Luo}, \bibinfo{person}{Yali Wang}, \bibinfo{person}{Limin Wang}, {and} \bibinfo{person}{Yu Qiao}.} \bibinfo{year}{2023}\natexlab{}.
\newblock \showarticletitle{Videochat: Chat-centric video understanding}.
\newblock \bibinfo{journal}{\emph{arXiv preprint arXiv:2305.06355}} (\bibinfo{year}{2023}).
\newblock


\bibitem[Li et~al\mbox{.}(2024a)]%
        {li2024self}
\bibfield{author}{\bibinfo{person}{Zhuo Li}, \bibinfo{person}{Yuhao Du}, \bibinfo{person}{Jinpeng Hu}, \bibinfo{person}{Xiang Wan}, {and} \bibinfo{person}{Anningzhe Gao}.} \bibinfo{year}{2024}\natexlab{a}.
\newblock \showarticletitle{Self-instructed derived prompt generation meets in-context learning: Unlocking new potential of black-box llms}.
\newblock \bibinfo{journal}{\emph{arXiv preprint arXiv:2409.01552}} (\bibinfo{year}{2024}).
\newblock


\bibitem[Li et~al\mbox{.}(2025a)]%
        {li2025add}
\bibfield{author}{\bibinfo{person}{Zhuo Li}, \bibinfo{person}{Yuhao Du}, \bibinfo{person}{Xiaoqi Jiao}, \bibinfo{person}{Yiwen Guo}, \bibinfo{person}{Yuege Feng}, \bibinfo{person}{Xiang Wan}, \bibinfo{person}{Anningzhe Gao}, {and} \bibinfo{person}{Jinpeng Hu}.} \bibinfo{year}{2025}\natexlab{a}.
\newblock \showarticletitle{Add-One-In: Incremental Sample Selection for Large Language Models via a Choice-Based Greedy Paradigm}.
\newblock \bibinfo{journal}{\emph{arXiv preprint arXiv:2503.02359}} (\bibinfo{year}{2025}).
\newblock


\bibitem[Li et~al\mbox{.}(2025b)]%
        {li2025prototype}
\bibfield{author}{\bibinfo{person}{Zhuo Li}, \bibinfo{person}{He Zhao}, \bibinfo{person}{Anningzhe Gao}, \bibinfo{person}{Dandan Guo}, \bibinfo{person}{Tsung-Hui Chang}, {and} \bibinfo{person}{Xiang Wan}.} \bibinfo{year}{2025}\natexlab{b}.
\newblock \showarticletitle{Prototype-oriented clean subset extraction for noisy long-tailed classification}.
\newblock \bibinfo{journal}{\emph{IEEE Transactions on Circuits and Systems for Video Technology}} (\bibinfo{year}{2025}).
\newblock


\bibitem[Lin et~al\mbox{.}(2023)]%
        {lin2023video}
\bibfield{author}{\bibinfo{person}{Bin Lin}, \bibinfo{person}{Bin Zhu}, \bibinfo{person}{Yang Ye}, \bibinfo{person}{Munan Ning}, \bibinfo{person}{Peng Jin}, {and} \bibinfo{person}{Li Yuan}.} \bibinfo{year}{2023}\natexlab{}.
\newblock \showarticletitle{Video-llava: Learning united visual representation by alignment before projection}.
\newblock \bibinfo{journal}{\emph{arXiv preprint arXiv:2311.10122}} (\bibinfo{year}{2023}).
\newblock


\bibitem[Liu et~al\mbox{.}(2024b)]%
        {liu2024visual}
\bibfield{author}{\bibinfo{person}{Haotian Liu}, \bibinfo{person}{Chunyuan Li}, \bibinfo{person}{Qingyang Wu}, {and} \bibinfo{person}{Yong~Jae Lee}.} \bibinfo{year}{2024}\natexlab{b}.
\newblock \showarticletitle{Visual instruction tuning}.
\newblock \bibinfo{journal}{\emph{Advances in neural information processing systems}}  \bibinfo{volume}{36} (\bibinfo{year}{2024}).
\newblock


\bibitem[Liu and Lapata(2019)]%
        {liu2019text}
\bibfield{author}{\bibinfo{person}{Yang Liu} {and} \bibinfo{person}{Mirella Lapata}.} \bibinfo{year}{2019}\natexlab{}.
\newblock \showarticletitle{Text Summarization with Pretrained Encoders}. In \bibinfo{booktitle}{\emph{Proceedings of the 2019 Conference on Empirical Methods in Natural Language Processing and the 9th International Joint Conference on Natural Language Processing (EMNLP-IJCNLP)}}. \bibinfo{pages}{3730--3740}.
\newblock


\bibitem[Liu et~al\mbox{.}(2024a)]%
        {liu2024oryx}
\bibfield{author}{\bibinfo{person}{Zuyan Liu}, \bibinfo{person}{Yuhao Dong}, \bibinfo{person}{Ziwei Liu}, \bibinfo{person}{Winston Hu}, \bibinfo{person}{Jiwen Lu}, {and} \bibinfo{person}{Yongming Rao}.} \bibinfo{year}{2024}\natexlab{a}.
\newblock \showarticletitle{Oryx MLLM: On-Demand Spatial-Temporal Understanding at Arbitrary Resolution}.
\newblock \bibinfo{journal}{\emph{arXiv preprint arXiv:2409.12961}} (\bibinfo{year}{2024}).
\newblock


\bibitem[Maaz et~al\mbox{.}(2023)]%
        {maaz2023video}
\bibfield{author}{\bibinfo{person}{Muhammad Maaz}, \bibinfo{person}{Hanoona Rasheed}, \bibinfo{person}{Salman Khan}, {and} \bibinfo{person}{Fahad~Shahbaz Khan}.} \bibinfo{year}{2023}\natexlab{}.
\newblock \showarticletitle{Video-chatgpt: Towards detailed video understanding via large vision and language models}.
\newblock \bibinfo{journal}{\emph{arXiv preprint arXiv:2306.05424}} (\bibinfo{year}{2023}).
\newblock


\bibitem[{Mixkit}(2023)]%
        {mixkit}
\bibfield{author}{\bibinfo{person}{{Mixkit}}.} \bibinfo{year}{2023}\natexlab{}.
\newblock \bibinfo{title}{{Mixkit}: Free Assets for Video, Music, and Sound Effects}.
\newblock
\newblock
\newblock
\shownote{\url{https://mixkit.co}}.


\bibitem[Morency et~al\mbox{.}(2011)]%
        {morency2011towards}
\bibfield{author}{\bibinfo{person}{Louis-Philippe Morency}, \bibinfo{person}{Rada Mihalcea}, {and} \bibinfo{person}{Payal Doshi}.} \bibinfo{year}{2011}\natexlab{}.
\newblock \showarticletitle{Towards multimodal sentiment analysis: Harvesting opinions from the web}. In \bibinfo{booktitle}{\emph{Proceedings of the 13th international conference on multimodal interfaces}}. \bibinfo{pages}{169--176}.
\newblock


\bibitem[Mundnich et~al\mbox{.}(2020)]%
        {mundnich2020tiles}
\bibfield{author}{\bibinfo{person}{Karel Mundnich}, \bibinfo{person}{Brandon~M Booth}, \bibinfo{person}{Michelle l’Hommedieu}, \bibinfo{person}{Tiantian Feng}, \bibinfo{person}{Benjamin Girault}, \bibinfo{person}{Justin L’hommedieu}, \bibinfo{person}{Mackenzie Wildman}, \bibinfo{person}{Sophia Skaaden}, \bibinfo{person}{Amrutha Nadarajan}, \bibinfo{person}{Jennifer~L Villatte}, {et~al\mbox{.}}} \bibinfo{year}{2020}\natexlab{}.
\newblock \showarticletitle{TILES-2018, a longitudinal physiologic and behavioral data set of hospital workers}.
\newblock \bibinfo{journal}{\emph{Scientific Data}} \bibinfo{volume}{7}, \bibinfo{number}{1} (\bibinfo{year}{2020}), \bibinfo{pages}{354}.
\newblock


\bibitem[{Pexels}(2023)]%
        {pexels}
\bibfield{author}{\bibinfo{person}{{Pexels}}.} \bibinfo{year}{2023}\natexlab{}.
\newblock \bibinfo{title}{{Pexels}: Free Stock Photos and Videos}.
\newblock
\newblock
\newblock
\shownote{\url{https://www.pexels.com}}.


\bibitem[Poria et~al\mbox{.}(2019)]%
        {poria2019meld}
\bibfield{author}{\bibinfo{person}{Soujanya Poria}, \bibinfo{person}{Devamanyu Hazarika}, \bibinfo{person}{Navonil Majumder}, \bibinfo{person}{Gautam Naik}, \bibinfo{person}{Erik Cambria}, {and} \bibinfo{person}{Rada Mihalcea}.} \bibinfo{year}{2019}\natexlab{}.
\newblock \showarticletitle{MELD: A Multimodal Multi-Party Dataset for Emotion Recognition in Conversations}. In \bibinfo{booktitle}{\emph{Proceedings of the 57th Annual Meeting of the Association for Computational Linguistics}}. \bibinfo{pages}{527--536}.
\newblock


\bibitem[Radford et~al\mbox{.}(2021)]%
        {radford2021learning}
\bibfield{author}{\bibinfo{person}{Alec Radford}, \bibinfo{person}{Jong~Wook Kim}, \bibinfo{person}{Chris Hallacy}, \bibinfo{person}{Aditya Ramesh}, \bibinfo{person}{Gabriel Goh}, \bibinfo{person}{Sandhini Agarwal}, \bibinfo{person}{Girish Sastry}, \bibinfo{person}{Amanda Askell}, \bibinfo{person}{Pamela Mishkin}, \bibinfo{person}{Jack Clark}, {et~al\mbox{.}}} \bibinfo{year}{2021}\natexlab{}.
\newblock \showarticletitle{Learning transferable visual models from natural language supervision}. In \bibinfo{booktitle}{\emph{International conference on machine learning}}. PMLR, \bibinfo{pages}{8748--8763}.
\newblock


\bibitem[Rashkin et~al\mbox{.}({[n.\,d.]})]%
        {rashkin1811towards}
\bibfield{author}{\bibinfo{person}{H Rashkin}, \bibinfo{person}{EM Smith}, \bibinfo{person}{M Li}, {and} \bibinfo{person}{YL Boureau}.} \bibinfo{year}{[n.\,d.]}\natexlab{}.
\newblock \showarticletitle{Towards empathetic open-domain conversation models: A new benchmark and dataset. arXiv 2018}.
\newblock \bibinfo{journal}{\emph{arXiv preprint arXiv:1811.00207}} (\bibinfo{year}{[n.\,d.]}).
\newblock


\bibitem[Shen et~al\mbox{.}(2020)]%
        {shen2020memor}
\bibfield{author}{\bibinfo{person}{Guangyao Shen}, \bibinfo{person}{Xin Wang}, \bibinfo{person}{Xuguang Duan}, \bibinfo{person}{Hongzhi Li}, {and} \bibinfo{person}{Wenwu Zhu}.} \bibinfo{year}{2020}\natexlab{}.
\newblock \showarticletitle{Memor: A dataset for multimodal emotion reasoning in videos}. In \bibinfo{booktitle}{\emph{Proceedings of the 28th ACM international conference on multimedia}}. \bibinfo{pages}{493--502}.
\newblock


\bibitem[Song et~al\mbox{.}(2023a)]%
        {song2023contextual}
\bibfield{author}{\bibinfo{person}{Peipei Song}, \bibinfo{person}{Dan Guo}, \bibinfo{person}{Jun Cheng}, {and} \bibinfo{person}{Meng Wang}.} \bibinfo{year}{2023}\natexlab{a}.
\newblock \showarticletitle{Contextual Attention Network for Emotional Video Captioning}.
\newblock \bibinfo{journal}{\emph{IEEE Transactions on Multimedia}}  \bibinfo{volume}{25} (\bibinfo{year}{2023}), \bibinfo{pages}{1858--1867}.
\newblock


\bibitem[Song et~al\mbox{.}(2024)]%
        {song2024emotional}
\bibfield{author}{\bibinfo{person}{Peipei Song}, \bibinfo{person}{Dan Guo}, \bibinfo{person}{Xun Yang}, \bibinfo{person}{Shengeng Tang}, {and} \bibinfo{person}{Meng Wang}.} \bibinfo{year}{2024}\natexlab{}.
\newblock \showarticletitle{Emotional Video Captioning With Vision-Based Emotion Interpretation Network}.
\newblock \bibinfo{journal}{\emph{IEEE Transactions on Image Processing}}  \bibinfo{volume}{33} (\bibinfo{year}{2024}), \bibinfo{pages}{1122--1135}.
\newblock


\bibitem[Song et~al\mbox{.}(2023b)]%
        {song2023emotion}
\bibfield{author}{\bibinfo{person}{Peipei Song}, \bibinfo{person}{Dan Guo}, \bibinfo{person}{Xun Yang}, \bibinfo{person}{Shengeng Tang}, \bibinfo{person}{Erkun Yang}, {and} \bibinfo{person}{Meng Wang}.} \bibinfo{year}{2023}\natexlab{b}.
\newblock \showarticletitle{Emotion-Prior Awareness Network for Emotional Video Captioning}. In \bibinfo{booktitle}{\emph{Proceedings of the 31st ACM International Conference on Multimedia}}. \bibinfo{pages}{589–600}.
\newblock


\bibitem[Stappen et~al\mbox{.}(2021)]%
        {stappen2021multimodal}
\bibfield{author}{\bibinfo{person}{Lukas Stappen}, \bibinfo{person}{Alice Baird}, \bibinfo{person}{Lea Schumann}, {and} \bibinfo{person}{Bj{\"o}rn Schuller}.} \bibinfo{year}{2021}\natexlab{}.
\newblock \showarticletitle{The multimodal sentiment analysis in car reviews (muse-car) dataset: Collection, insights and improvements}.
\newblock \bibinfo{journal}{\emph{IEEE Transactions on Affective Computing}} \bibinfo{volume}{14}, \bibinfo{number}{2} (\bibinfo{year}{2021}), \bibinfo{pages}{1334--1350}.
\newblock


\bibitem[Touvron et~al\mbox{.}(2023)]%
        {touvron2023llama}
\bibfield{author}{\bibinfo{person}{Hugo Touvron}, \bibinfo{person}{Thibaut Lavril}, \bibinfo{person}{Gautier Izacard}, \bibinfo{person}{Xavier Martinet}, \bibinfo{person}{Marie-Anne Lachaux}, \bibinfo{person}{Timoth{\'e}e Lacroix}, \bibinfo{person}{Baptiste Rozi{\`e}re}, \bibinfo{person}{Naman Goyal}, \bibinfo{person}{Eric Hambro}, \bibinfo{person}{Faisal Azhar}, {et~al\mbox{.}}} \bibinfo{year}{2023}\natexlab{}.
\newblock \showarticletitle{Llama: Open and efficient foundation language models}.
\newblock \bibinfo{journal}{\emph{arXiv preprint arXiv:2302.13971}} (\bibinfo{year}{2023}).
\newblock


\bibitem[Turcan and McKeown(2019)]%
        {turcan2019dreaddit}
\bibfield{author}{\bibinfo{person}{Elsbeth Turcan} {and} \bibinfo{person}{Kathleen McKeown}.} \bibinfo{year}{2019}\natexlab{}.
\newblock \showarticletitle{Dreaddit: A reddit dataset for stress analysis in social media}.
\newblock \bibinfo{journal}{\emph{arXiv preprint arXiv:1911.00133}} (\bibinfo{year}{2019}).
\newblock


\bibitem[Vaswani et~al\mbox{.}(2017)]%
        {vaswani2017attention}
\bibfield{author}{\bibinfo{person}{Ashish Vaswani}, \bibinfo{person}{Noam Shazeer}, \bibinfo{person}{Niki Parmar}, \bibinfo{person}{Jakob Uszkoreit}, \bibinfo{person}{Llion Jones}, \bibinfo{person}{Aidan~N Gomez}, \bibinfo{person}{{\L}ukasz Kaiser}, {and} \bibinfo{person}{Illia Polosukhin}.} \bibinfo{year}{2017}\natexlab{}.
\newblock \showarticletitle{Attention is all you need}.
\newblock \bibinfo{journal}{\emph{Advances in neural information processing systems}}  \bibinfo{volume}{30} (\bibinfo{year}{2017}).
\newblock


\bibitem[Wang et~al\mbox{.}(2024)]%
        {wang2024qwen2}
\bibfield{author}{\bibinfo{person}{Peng Wang}, \bibinfo{person}{Shuai Bai}, \bibinfo{person}{Sinan Tan}, \bibinfo{person}{Shijie Wang}, \bibinfo{person}{Zhihao Fan}, \bibinfo{person}{Jinze Bai}, \bibinfo{person}{Keqin Chen}, \bibinfo{person}{Xuejing Liu}, \bibinfo{person}{Jialin Wang}, \bibinfo{person}{Wenbin Ge}, {et~al\mbox{.}}} \bibinfo{year}{2024}\natexlab{}.
\newblock \showarticletitle{Qwen2-VL: Enhancing Vision-Language Model's Perception of the World at Any Resolution}.
\newblock \bibinfo{journal}{\emph{arXiv preprint arXiv:2409.12191}} (\bibinfo{year}{2024}).
\newblock


\bibitem[Wei et~al\mbox{.}(2021)]%
        {wei2021finetuned}
\bibfield{author}{\bibinfo{person}{Jason Wei}, \bibinfo{person}{Maarten Bosma}, \bibinfo{person}{Vincent~Y Zhao}, \bibinfo{person}{Kelvin Guu}, \bibinfo{person}{Adams~Wei Yu}, \bibinfo{person}{Brian Lester}, \bibinfo{person}{Nan Du}, \bibinfo{person}{Andrew~M Dai}, {and} \bibinfo{person}{Quoc~V Le}.} \bibinfo{year}{2021}\natexlab{}.
\newblock \showarticletitle{Finetuned language models are zero-shot learners}.
\newblock \bibinfo{journal}{\emph{arXiv preprint arXiv:2109.01652}} (\bibinfo{year}{2021}).
\newblock


\bibitem[Wei et~al\mbox{.}(2022)]%
        {wei2022chain}
\bibfield{author}{\bibinfo{person}{Jason Wei}, \bibinfo{person}{Xuezhi Wang}, \bibinfo{person}{Dale Schuurmans}, \bibinfo{person}{Maarten Bosma}, \bibinfo{person}{Fei Xia}, \bibinfo{person}{Ed Chi}, \bibinfo{person}{Quoc~V Le}, \bibinfo{person}{Denny Zhou}, {et~al\mbox{.}}} \bibinfo{year}{2022}\natexlab{}.
\newblock \showarticletitle{Chain-of-thought prompting elicits reasoning in large language models}.
\newblock \bibinfo{journal}{\emph{Advances in neural information processing systems}}  \bibinfo{volume}{35} (\bibinfo{year}{2022}), \bibinfo{pages}{24824--24837}.
\newblock


\bibitem[Wilf et~al\mbox{.}(2023)]%
        {siq2}
\bibfield{author}{\bibinfo{person}{Alex Wilf}, \bibinfo{person}{Leena Mathur}, \bibinfo{person}{Sheryl Mathew}, \bibinfo{person}{Claire Ko}, \bibinfo{person}{Youssouf Kebe}, \bibinfo{person}{Paul~Pu Liang}, {and} \bibinfo{person}{Louis-Philippe Morency}.} \bibinfo{year}{2023}\natexlab{}.
\newblock \bibinfo{title}{Social-IQ 2.0 Challenge: Benchmarking Multimodal Social Understanding}.
\newblock \bibinfo{howpublished}{\url{https://github.com/abwilf/Social-IQ-2.0-Challenge}}.
\newblock


\bibitem[W{\"o}llmer et~al\mbox{.}(2013)]%
        {wollmer2013youtube}
\bibfield{author}{\bibinfo{person}{Martin W{\"o}llmer}, \bibinfo{person}{Felix Weninger}, \bibinfo{person}{Tobias Knaup}, \bibinfo{person}{Bj{\"o}rn Schuller}, \bibinfo{person}{Congkai Sun}, \bibinfo{person}{Kenji Sagae}, {and} \bibinfo{person}{Louis-Philippe Morency}.} \bibinfo{year}{2013}\natexlab{}.
\newblock \showarticletitle{Youtube movie reviews: Sentiment analysis in an audio-visual context}.
\newblock \bibinfo{journal}{\emph{IEEE Intelligent Systems}} \bibinfo{volume}{28}, \bibinfo{number}{3} (\bibinfo{year}{2013}), \bibinfo{pages}{46--53}.
\newblock


\bibitem[Wu et~al\mbox{.}(2024)]%
        {wu2024autogen}
\bibfield{author}{\bibinfo{person}{Qingyun Wu}, \bibinfo{person}{Gagan Bansal}, \bibinfo{person}{Jieyu Zhang}, \bibinfo{person}{Yiran Wu}, \bibinfo{person}{Beibin Li}, \bibinfo{person}{Erkang Zhu}, \bibinfo{person}{Li Jiang}, \bibinfo{person}{Xiaoyun Zhang}, \bibinfo{person}{Shaokun Zhang}, \bibinfo{person}{Jiale Liu}, {et~al\mbox{.}}} \bibinfo{year}{2024}\natexlab{}.
\newblock \showarticletitle{Autogen: Enabling next-gen LLM applications via multi-agent conversations}. In \bibinfo{booktitle}{\emph{First Conference on Language Modeling}}.
\newblock


\bibitem[Xu et~al\mbox{.}(2022)]%
        {xu2022mcpr}
\bibfield{author}{\bibinfo{person}{Carol Xu}, \bibinfo{person}{Xuan Luo}, {and} \bibinfo{person}{Dan Wang}.} \bibinfo{year}{2022}\natexlab{}.
\newblock \showarticletitle{MCPR: a Chinese product review dataset for multimodal aspect-based sentiment analysis}. In \bibinfo{booktitle}{\emph{International Conference on Cognitive Computing}}. Springer, \bibinfo{pages}{83--90}.
\newblock


\bibitem[Xu et~al\mbox{.}(2023)]%
        {xu2023wizardlmempoweringlargelanguage}
\bibfield{author}{\bibinfo{person}{Can Xu}, \bibinfo{person}{Qingfeng Sun}, \bibinfo{person}{Kai Zheng}, \bibinfo{person}{Xiubo Geng}, \bibinfo{person}{Pu Zhao}, \bibinfo{person}{Jiazhan Feng}, \bibinfo{person}{Chongyang Tao}, {and} \bibinfo{person}{Daxin Jiang}.} \bibinfo{year}{2023}\natexlab{}.
\newblock \bibinfo{title}{WizardLM: Empowering Large Language Models to Follow Complex Instructions}.
\newblock
\newblock
\showeprint[arxiv]{2304.12244}~[cs.CL]
\urldef\tempurl%
\url{https://arxiv.org/abs/2304.12244}
\showURL{%
\tempurl}


\bibitem[Xu et~al\mbox{.}(2019)]%
        {xu2019multi}
\bibfield{author}{\bibinfo{person}{Nan Xu}, \bibinfo{person}{Wenji Mao}, {and} \bibinfo{person}{Guandan Chen}.} \bibinfo{year}{2019}\natexlab{}.
\newblock \showarticletitle{Multi-interactive memory network for aspect based multimodal sentiment analysis}. In \bibinfo{booktitle}{\emph{Proceedings of the AAAI conference on artificial intelligence}}, Vol.~\bibinfo{volume}{33}. \bibinfo{pages}{371--378}.
\newblock


\bibitem[Yan et~al\mbox{.}(2024)]%
        {yan2024list}
\bibfield{author}{\bibinfo{person}{An Yan}, \bibinfo{person}{Zhengyuan Yang}, \bibinfo{person}{Junda Wu}, \bibinfo{person}{Wanrong Zhu}, \bibinfo{person}{Jianwei Yang}, \bibinfo{person}{Linjie Li}, \bibinfo{person}{Kevin Lin}, \bibinfo{person}{Jianfeng Wang}, \bibinfo{person}{Julian McAuley}, \bibinfo{person}{Jianfeng Gao}, {et~al\mbox{.}}} \bibinfo{year}{2024}\natexlab{}.
\newblock \showarticletitle{List Items One by One: A New Data Source and Learning Paradigm for Multimodal LLMs}.
\newblock \bibinfo{journal}{\emph{arXiv preprint arXiv:2404.16375}} (\bibinfo{year}{2024}).
\newblock


\bibitem[Yasunaga et~al\mbox{.}(2024)]%
        {yasunaga2024large}
\bibfield{author}{\bibinfo{person}{Michihiro Yasunaga}, \bibinfo{person}{Xinyun Chen}, \bibinfo{person}{Yujia Li}, \bibinfo{person}{Panupong Pasupat}, \bibinfo{person}{Jure Leskovec}, \bibinfo{person}{Percy Liang}, \bibinfo{person}{Ed~H. Chi}, {and} \bibinfo{person}{Denny Zhou}.} \bibinfo{year}{2024}\natexlab{}.
\newblock \showarticletitle{Large Language Models as Analogical Reasoners}. In \bibinfo{booktitle}{\emph{The Twelfth International Conference on Learning Representations}}.
\newblock
\urldef\tempurl%
\url{https://openreview.net/forum?id=AgDICX1h50}
\showURL{%
\tempurl}


\bibitem[Zadeh et~al\mbox{.}(2020)]%
        {zadeh2020cmu}
\bibfield{author}{\bibinfo{person}{Amir Zadeh}, \bibinfo{person}{Yan~Sheng Cao}, \bibinfo{person}{Simon Hessner}, \bibinfo{person}{Paul~Pu Liang}, \bibinfo{person}{Soujanya Poria}, {and} \bibinfo{person}{Louis-Philippe Morency}.} \bibinfo{year}{2020}\natexlab{}.
\newblock \showarticletitle{CMU-MOSEAS: A multimodal language dataset for Spanish, Portuguese, German and French}. In \bibinfo{booktitle}{\emph{Proceedings of the Conference on Empirical Methods in Natural Language Processing. Conference on Empirical Methods in Natural Language Processing}}, Vol.~\bibinfo{volume}{2020}. NIH Public Access, \bibinfo{pages}{1801}.
\newblock


\bibitem[Zadeh et~al\mbox{.}(2019)]%
        {zadeh2019social}
\bibfield{author}{\bibinfo{person}{Amir Zadeh}, \bibinfo{person}{Michael Chan}, \bibinfo{person}{Paul~Pu Liang}, \bibinfo{person}{Edmund Tong}, {and} \bibinfo{person}{Louis-Philippe Morency}.} \bibinfo{year}{2019}\natexlab{}.
\newblock \showarticletitle{Social-iq: A question answering benchmark for artificial social intelligence}. In \bibinfo{booktitle}{\emph{Proceedings of the IEEE/CVF Conference on Computer Vision and Pattern Recognition}}. \bibinfo{pages}{8807--8817}.
\newblock


\bibitem[Zadeh et~al\mbox{.}(2016)]%
        {zadeh2016multimodal}
\bibfield{author}{\bibinfo{person}{Amir Zadeh}, \bibinfo{person}{Rowan Zellers}, \bibinfo{person}{Eli Pincus}, {and} \bibinfo{person}{Louis-Philippe Morency}.} \bibinfo{year}{2016}\natexlab{}.
\newblock \showarticletitle{Multimodal sentiment intensity analysis in videos: Facial gestures and verbal messages}.
\newblock \bibinfo{journal}{\emph{IEEE Intelligent Systems}} \bibinfo{volume}{31}, \bibinfo{number}{6} (\bibinfo{year}{2016}), \bibinfo{pages}{82--88}.
\newblock


\bibitem[Zadeh et~al\mbox{.}(2018)]%
        {zadeh2018multimodal}
\bibfield{author}{\bibinfo{person}{AmirAli~Bagher Zadeh}, \bibinfo{person}{Paul~Pu Liang}, \bibinfo{person}{Soujanya Poria}, \bibinfo{person}{Erik Cambria}, {and} \bibinfo{person}{Louis-Philippe Morency}.} \bibinfo{year}{2018}\natexlab{}.
\newblock \showarticletitle{Multimodal language analysis in the wild: Cmu-mosei dataset and interpretable dynamic fusion graph}. In \bibinfo{booktitle}{\emph{Proceedings of the 56th Annual Meeting of the Association for Computational Linguistics (Volume 1: Long Papers)}}. \bibinfo{pages}{2236--2246}.
\newblock


\bibitem[Zhang et~al\mbox{.}(2023)]%
        {zhang2023llavar}
\bibfield{author}{\bibinfo{person}{Yanzhe Zhang}, \bibinfo{person}{Ruiyi Zhang}, \bibinfo{person}{Jiuxiang Gu}, \bibinfo{person}{Yufan Zhou}, \bibinfo{person}{Nedim Lipka}, \bibinfo{person}{Diyi Yang}, {and} \bibinfo{person}{Tong Sun}.} \bibinfo{year}{2023}\natexlab{}.
\newblock \showarticletitle{Llavar: Enhanced visual instruction tuning for text-rich image understanding}.
\newblock \bibinfo{journal}{\emph{arXiv preprint arXiv:2306.17107}} (\bibinfo{year}{2023}).
\newblock


\bibitem[Zhao et~al\mbox{.}(2023)]%
        {zhao2023svit}
\bibfield{author}{\bibinfo{person}{Bo Zhao}, \bibinfo{person}{Boya Wu}, \bibinfo{person}{Muyang He}, {and} \bibinfo{person}{Tiejun Huang}.} \bibinfo{year}{2023}\natexlab{}.
\newblock \showarticletitle{Svit: Scaling up visual instruction tuning}.
\newblock \bibinfo{journal}{\emph{arXiv preprint arXiv:2307.04087}} (\bibinfo{year}{2023}).
\newblock


\bibitem[Zheng et~al\mbox{.}(2024)]%
        {zheng2024take}
\bibfield{author}{\bibinfo{person}{Huaixiu~Steven Zheng}, \bibinfo{person}{Swaroop Mishra}, \bibinfo{person}{Xinyun Chen}, \bibinfo{person}{Heng-Tze Cheng}, \bibinfo{person}{Ed~H. Chi}, \bibinfo{person}{Quoc~V Le}, {and} \bibinfo{person}{Denny Zhou}.} \bibinfo{year}{2024}\natexlab{}.
\newblock \showarticletitle{Take a Step Back: Evoking Reasoning via Abstraction in Large Language Models}. In \bibinfo{booktitle}{\emph{The Twelfth International Conference on Learning Representations}}.
\newblock
\urldef\tempurl%
\url{https://openreview.net/forum?id=3bq3jsvcQ1}
\showURL{%
\tempurl}


\bibitem[Zhu et~al\mbox{.}(2023)]%
        {zhu2023minigpt}
\bibfield{author}{\bibinfo{person}{Deyao Zhu}, \bibinfo{person}{Jun Chen}, \bibinfo{person}{Xiaoqian Shen}, \bibinfo{person}{Xiang Li}, {and} \bibinfo{person}{Mohamed Elhoseiny}.} \bibinfo{year}{2023}\natexlab{}.
\newblock \showarticletitle{Minigpt-4: Enhancing vision-language understanding with advanced large language models}.
\newblock \bibinfo{journal}{\emph{arXiv preprint arXiv:2304.10592}} (\bibinfo{year}{2023}).
\newblock


\end{thebibliography}


\end{document}